\documentclass[runningheads]{llncs}

\usepackage{eccv}

\usepackage{eccvabbrv}

\usepackage{graphicx}
\usepackage{booktabs}
\usepackage{multirow}
\usepackage{array}
\usepackage{tabularx}
\usepackage[accsupp]{axessibility}  

\usepackage[breaklinks,colorlinks]{hyperref}

\usepackage{orcidlink}

\begin{document}

\title{See, Remember, Explore: A Benchmark and Baselines for Streaming Spatial Reasoning} 

\titlerunning{See, Remember, Explore}


\authorrunning{W.~Author et al.}


\author{Yuxi Wei\inst{1} \and
Wei Huang\inst{1} \and
Qirui Chen\inst{3} \and
Lu Hou\inst{2}\thanks{Corresponding author} \and
Xiaojuan Qi\inst{1}\textsuperscript{*}}

\institute{The University of Hong Kong \and
Yinwang Intelligent Technology Co. Ltd. \and
Shanghai Jiao Tong University\\
\email{wyx3590236732@sjtu.edu.cn}\\
\url{https://vfishc.github.io/s3-bench}}

\maketitle

\begin{abstract}
Spatial understanding is fundamental for embodied agents, yet most spatial VLMs and benchmarks remain offline—evaluating post-hoc QA over pre-recorded inputs and overlooking two crucial deployment-critical requirements: long-horizon streaming inference and active perception when the current view is insufficient. To address this gap, we introduce $S^3$-Bench, a benchmark suite for streaming spatial question answering with active exploration, where queries are temporally grounded to specific timestamps and must be answered using only observations available up to that moment. $S^3$-Bench adopts a dual-domain design, combining a scalable simulator with controllable trajectories and exploration actions, and real-world streaming videos that capture practical sensing artifacts for rigorous generalization evaluation. Overall, it spans 10K+ scenes and 26K+ trajectories, with dedicated training ($S^3$-Train) and evaluation ($S^3$-Eval) splits. We further propose AMF-VLM, which supports streaming spatial reasoning under bounded computing via (i) memory folding, which compresses long-horizon observations into compact structured memory, and (ii) active exploration, which outputs explicit actions (\textit{e.g.} move/rotate/scan) to acquire missing evidence before answering.  Extensive experiments demonstrate that, compared to models using identical training data, our approach yields improvements of 8.8\% and 13.3\% on the simulated and real splits of $S^3$-Eval, respectively, while maintaining competitive transferability to standard spatial benchmarks. 

\keywords{Spatial Understanding \and Embodied AI \and Active Vision \and Streaming Perception}
\end{abstract}
\section{Introduction}
\begin{figure}[t]
    \centering
    \includegraphics[width=0.85\linewidth]{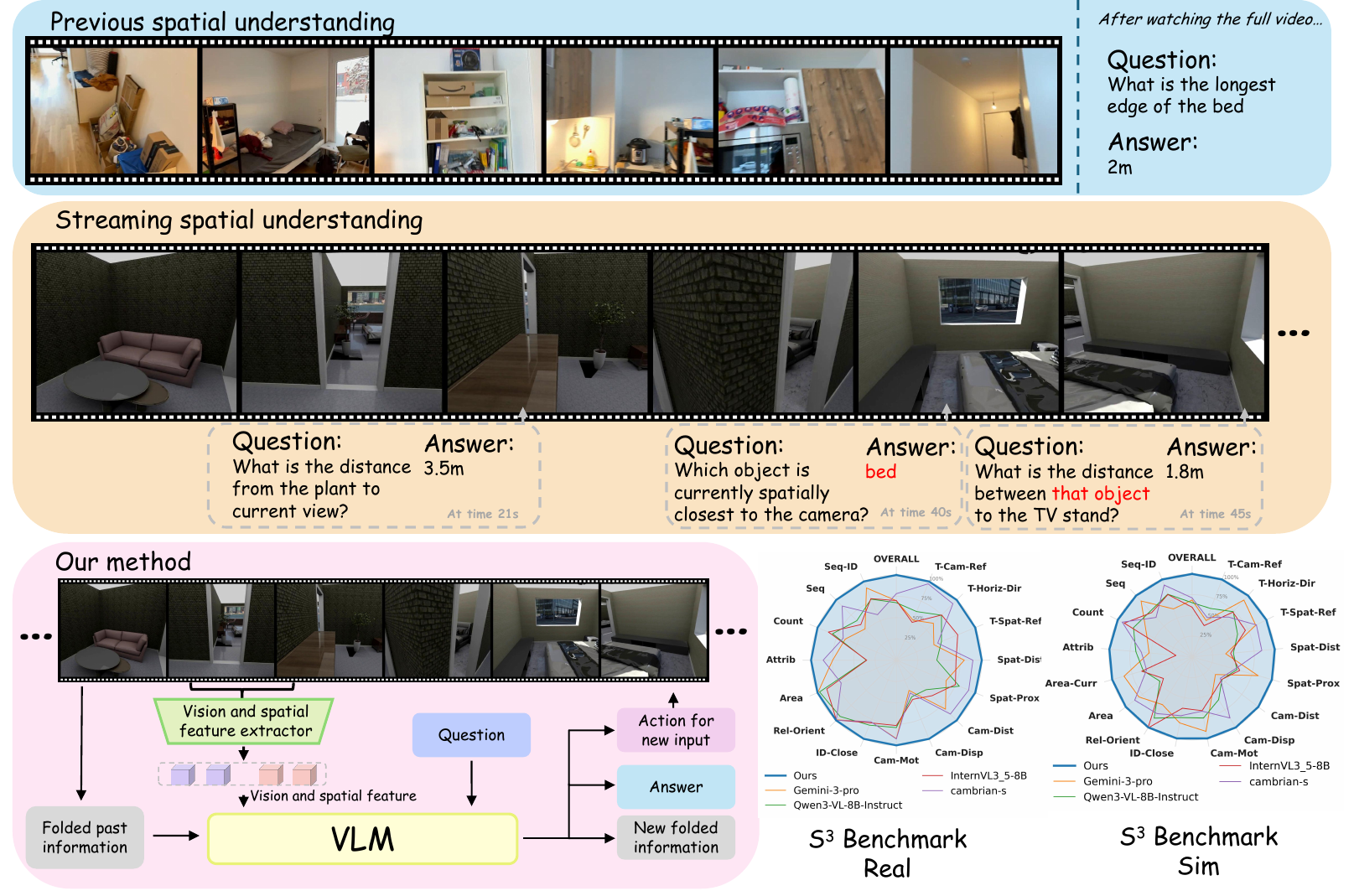}
    \vspace{-1.5mm}
    \caption{\small \textbf{Overview of the $S^3$-Bench and AMF-VLM.} 
        Unlike traditional offline methods that answer questions post-video, our $S^3$-Bench targets streaming spatial understanding with timestamp-specific, ego-centric, and temporally dependent queries. The figure outlines the AMF-VLM pipeline and presents its performance comparison with baselines on $S^3$-Eval.}
    \label{fig:teaser}

    \vspace{-5mm}
\end{figure}

Spatial understanding -- reasoning about objects, geometry, and their relationships in an environment -- is a prerequisite for embodied intelligence \cite{majumdar2024openeqa, azuma2022scanqa, cheng2024spatialrgpt, song2025robospatial, team2025robobrain}. An agent that can \textit{navigate, plan, and manipulate} must go beyond recognizing ``what is where'' in a single frame: it must infer distances, orientations, containment, and room-level structure, and maintain these beliefs as the viewpoint changes over time \cite{liu2023visual, cheng2024spatialrgpt, ma20253dsrbench}. In the vision-language era, VLMs have shown encouraging progress on spatial question answering \cite{xu2024vlm, yang2025cambrian, fan2025vlm}. Yet much of this progress is established under an \textbf{offline} assumption: the model is given a pre-recorded video (or a curated subset of frames) and answers questions after the full sequence is available, effectively treating space as something to be retrospectively analyzed rather than continuously tracked in the moment \cite{lin2025mmsi, yang2025thinking, li2024mvbench}.

However, real embodied agents do not operate offline. They perceive the world as a stream, under partial observability, and often must decide what to look at next to resolve ambiguity \cite{zhao2025vla, mavrogiannis2023core, yamada2023evaluating, lin2024streamingbench, qian2024streaming, zhao2023streaming}. This mismatch exposes two fundamental limitations of existing approaches. \textbf{(a) Offline perception and passive sensing}. In deployment, observations arrive continuously and the history grows without bound; embodied agents are frequently required to reason about the scene for planning tasks at specific moments, necessitating active exploration to retrieve occluded or unobserved information. Nevertheless, fixed, passive trajectories and post-hoc QA protocols in previous offline settings \cite{lin2025mmsi, yang2025thinking, fu2024blink} under-test these behaviors that matter for embodied operation. \textbf{(b) Offline-centric VLM design}. Most spatial VLMs are optimized for discrete, offline inputs (\textit{e.g.}, curated frame sets or completed reconstructions) \cite{zhao2025spacemind, qi2025gpt4scene, wu2025spatial, wang2025ross3d, li2025imagine, lee2025perspective}. When directly applied to streaming inputs, they encounter structural failure modes: the KV cache grows with every new frame, making long episodes computationally impractical, while long-horizon spatial information is gradually diluted, forgotten, or overwritten-- undermining persistent spatial understanding.

To close the gap between previous offline evaluation and embodied deployment, we introduce $S^3$-Bench, a benchmark suite comprising both training ($S^3$-Train) and evaluation ($S^3$-Eval) data for streaming spatial question answering with active exploration (Fig.~\ref{fig:teaser}). $S^3$-Bench adopts a dual-domain design: a simulator that supports free trajectory generation and controllable exploration actions (move/rotate/scan), enabling targeted tests without sufficient information such as occlusions and long-horizon navigation; and real-world streaming videos that capture practical sensing artifacts (e.g. blur, lighting changes, clutter, and imperfect motion) to evaluate generalization. Together, these domains balance controlled scalability with realistic transfer. Crucially, $S^3$-Bench enforces temporally grounded evaluation: each query is anchored to a specific timestamp and must be answered using only observations available up to that moment, faithfully reflecting the causal constraints faced by embodied agents. In total, $S^3$-Bench spans over 10{,}000 scenes and more than 26{,}000 diverse camera trajectories. From this collection, we manually refine approximately 1{,}000 scenes and 20{,}000 QA pairs to construct $S^3$-Eval, while the remaining data forms $S^3$-Train (about 600K examples) to support learning for streaming spatial understanding.

Complementing the benchmark, we propose \textbf{AMF-VLM}, an Active Memory-Folding vision-language model for streaming spatial reasoning under bounded compute (Fig. \ref{fig:teaser}). AMF-VLM mainly consists of two novel modules. \textbf{Memory Folding} maintains a compact, persistent world state by periodically summarizing the growing stream into structured memory units-- capturing entities, attributes, and spatial relations (\textit{e.g.} object identities, relative positions, and landmark cues)-- and updating them as new evidence arrives. This ``folding'' operation controls the context length, mitigates KV-cache blow-up, and reduces long-horizon forgetting while retaining spatially decisive information needed for downstream queries. \textbf{Active Exploration} complements memory by handling insufficiency of evidence: given a query and the current memory state, AMF-VLM estimates what information is missing (\textit{e.g.} an occluded object or an unresolved relation) and emits explicit action proposals (\textit{e.g.} rotate to scan, move closer, step left/right) to acquire targeted observations. The newly gathered views are then integrated back into observation, after which the model produces a final, evidence-grounded answer based on the updated stream state. Our method consistently outperforms existing approaches on both the Sim and Real subsets of the $S^3$-Eval (Fig. \ref{fig:teaser}), even when other methods utilize the same training data.

In summary, the main contributions of this work are as follows:
\begin{itemize}
 \item We formalize streaming spatial understanding for VLMs, aligning evaluation and modeling with real embodied-agent deployment constraints (online perception, temporally grounded queries and long-horizon context). 
\item We introduce $S^3$-Bench, a comprehensive benchmark including training and evaluation data for streaming spatial QA and active exploration, spanning both simulator and real-world data to jointly support controllable exploration and generalization evaluation.
\item We propose AMF-VLM, featuring memory folding to control long-sequence costs and reduce forgetting, and an actionable exploration interface to acquire missing evidence when observations are insufficient.
\item Extensive evaluations demonstrate strong performance on $S^3$-Bench evaluation and competitive transfer to standard spatial benchmarks under streaming-ready inference.

\end{itemize}

\section{Related Work}

\noindent\textbf{Benchmarks for Spatial Understanding.}
Spatial VLMs are supported by benchmarks like ScanQA \cite{azuma2022scanqa}, VSIBench \cite{yang2025thinking} and some others \cite{ma20253dsrbench, lin2025mmsi, yang2025mmsi, xu2025spatialbench, tian2025nuscenes, du2024embspatial, yu2025far, stogiannidis2025mind}, which primarily focus on offline QA. These benchmarks lack temporal grounded annotations for continuous inputs, and their reliance on static recordings limits scene diversity while precluding free-trajectory generation. Consequently, they cannot adequately test active exploration. By combining interactive simulators with real-world data and temporal grounded labeling, $S^3$-Bench supports both streaming spatial QA and unconstrained active exploration.

\vspace{0.05in}\noindent\textbf{Spatial Vision-Language Models.}
Recent advancements in Vision-Language Models (VLMs) have demonstrated notable progress in spatial understanding \cite{chen2024spatialvlm, hu2025g, gholami2025spatial, qu2025spatialvla, cai2025spatialbot, zhang2025spatial, yang2025cambrian, fan2025vlm}. But they follow an offline paradigm, processing fixed frames or pre-reconstructed 3D point clouds. This offline assumption restricts their applicability in real-world embodied tasks, which require continuous, streaming perception for timely interaction and reasoning.

\vspace{0.05in}\noindent\textbf{Memory Mechanisms.}
While memory mechanisms like \cite{sun2025scaling, su2026u, shao2025foldact, hu2025memory, zhang2025memory, xu2025mem, chhikara2025mem0, yan2025memory} are well-established in LLMs for long sequences, existing VLMs lack dedicated architectures to prevent forgetting in continuous spatial environments. To address this, we introduce an online memory mechanism tailored for streaming spatial understanding. By dynamically integrating folded scene descriptions into context, our approach mitigates historical forgetting and sustains long-horizon context under streaming-ready inference.

\vspace{0.05in}\noindent\textbf{Active Visual Exploration.}
Active vision enables agents to autonomously control sensors or generation model to acquire information \cite{chuang2025active, xiao2025ava, yu2025thinking, evans2024bad, yang2025mindjourney, chen2026spatial}. Although some VLMs incorporate exploration capabilities \cite{zhao2026cov, zhang2026think3d}, they typically operate under non-streaming conditions, diverging from the continuous reality of physical entities. Furthermore, limited environment interactivity often constrains broad exploration. Our approach aligns active exploration with the streaming process, enabling continuous and autonomous information seeking.
\section{$S^3$-Bench}

\begin{figure}[t]
    \centering
    \includegraphics[width=0.95\linewidth]{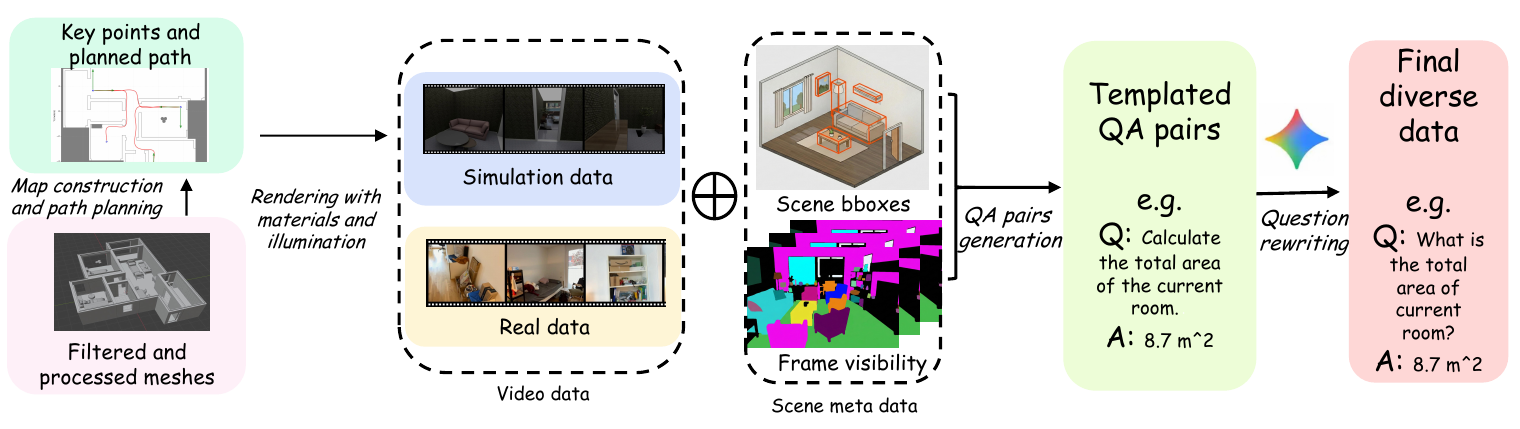}
    \caption{\small \textbf{$S^3$ Benchmark data construction pipeline.} Simulated digital assets are filtered and rendered into videos via geometry-guided path planning. Next, object bounding boxes and frame-level visibility are extracted to generate template-based QA pairs, which are then rewritten for linguistic diversity.}
    \label{fig:data_curation}
    \vspace{-3mm}
\end{figure}

To systematically evaluate spatial understanding under continuous visual inputs and active exploration, we introduce the Streaming Spatial Search Benchmark ($S^3$-Bench). The dataset is constructed across both simulated and real-world domains to ensure diverse, comprehensive evaluation and real-world generalization. 
Simulated data provides diverse scene layouts and flexible trajectory generation strategies, enabling precise acquisition of scene information and supporting active exploration environments. Real-world data more closely reflect actual usage distributions and better align with practical application scenarios. Approximately 1,000 scenes and 20,000 QA pairs are selected from the collection to construct the $S^3$-Eval following human refinement. The remaining part of the collection forms the $S^3$-Train 600K dataset. Fig. \ref{fig:data_curation} illustrates the construction pipeline. The pipeline collects videos and associated meta-information, specifically, object bounding boxes and frame-level object visibility, from both simulated and real-world environments. Based on these data, QA pairs are initially generated using templates and then rewritten via LLM to improve diversity.

\subsection{Simulated Environment Construction}
The simulated split of $S^3$-Bench provides more than 6K unique complex scene layouts and 20K camera trajectories, ensuring the diversity and scale of the data source. Additionally, the simulator offers freely explorable environments to support both training and testing for active exploration. The simulation pipeline first constructs and filters digital assets and then generates keypoints for each scene based on geometric properties. Afterward, it plans trajectories and renders the final videos alongside meta-information, including specifically generated samples tailored for active exploration.

\vspace{0.05in}\noindent\textbf{Asset Curation and Scene Layout.}  As shown in Fig.\ref{fig:data_curation},  firstly we source the foundational 3D assets and scene layouts from the 3D-FRONT dataset \cite{fu20213d}. Following a rigorous data cleaning process to eliminate topological anomalies and low-quality models, we compile over 6,000 distinct indoor scene layouts encompassing tens of thousands of individual rooms. To minimize the domain gap between simulation and reality, we significantly augment the visual fidelity of the scenes. This is achieved by randomly applying high-resolution wall and floor textures, alongside realistic environmental lighting models, sourced from extensive online repositories. These components enhance the visual fidelity and diversity of the environments, ultimately resulting in over 6K complete, composite, and renderable scenes.

\vspace{0.05in}\noindent\textbf{Complex Trajectory Generation.}  The prepared 3D assets are rendered using the specified camera paths to generate the final data. Unlike existing dataset from simulators \cite{brown2025sims, zhang2025spatial, kolve2017ai2, wang2024grutopia, li2023behavior} that predominantly rely on simplified random walks or basic shortest-path algorithms, we implement a systematic and geometry-aware trajectory generation pipeline. For each scene, we first extract the navigable mesh and identify physical obstacles. Within each discrete room, candidate waypoints are strategically sampled at key geometric locations, specifically room corners and the centers of maximum inscribed circles, or some targeted objects. At each keypoint, the camera stops to execute a sweeping motion. To plan the trajectory, we randomly select a subset of these keypoints formulated to maximize the overall spatial coverage of the scene. The final continuous trajectories are generated by routing through these keypoints using the A* search algorithm, followed by B-spline interpolation to ensure physical smoothness and kinematic realism. This methodology yields extensive, diverse, and complex scan trajectories. Utilizing the centers of large inscribed circles and wide-FoV corners maximizes visual information density, mimicking human observation logic. Subsequently, A* search and smoothing operations generate short, smooth trajectories that reflect realistic physical motion. Randomizing keypoints during planning guaranties data diversity, resulting in a robust set of high-quality, varied camera paths for rendering.

\begin{figure}[t]
    \centering
    \includegraphics[width=0.85\linewidth]{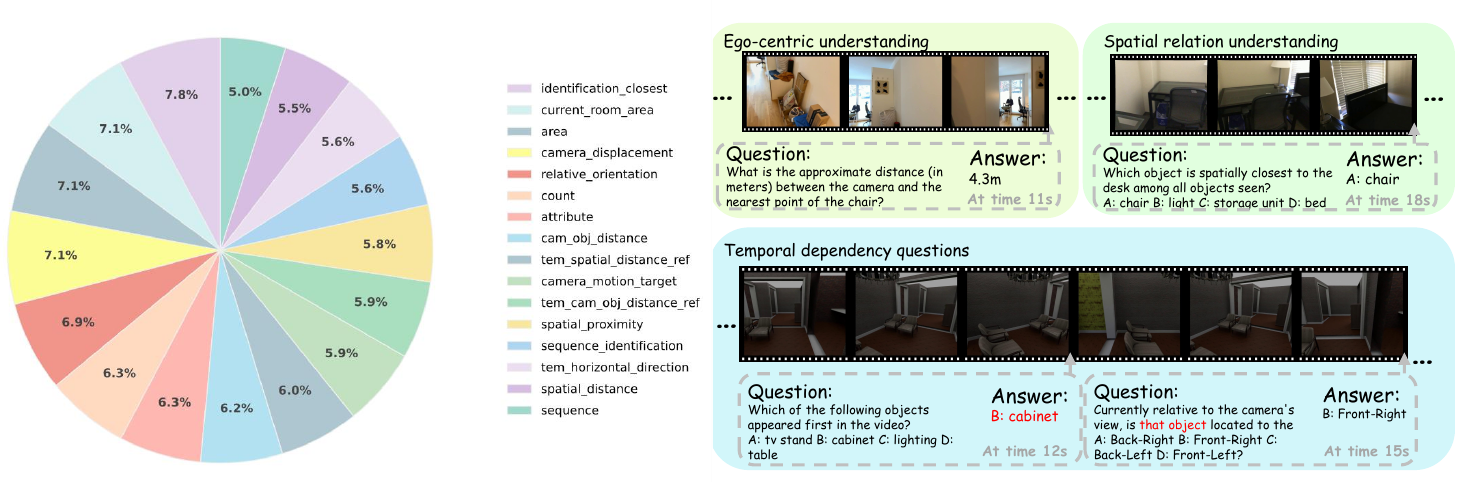}
    \vspace{-1.5mm}
    \caption{\small \textbf{Distribution and examples of QA categories in the $S^3$-Bench.}
This figure illustrates the statistical distribution of different QA categories alongside representative examples. The benchmark primarily comprises three question types: (1) \textit{Ego-centric}, requiring understanding based on the current camera position; (2) \textit{Spatial relation}, assessing the comprehension of the scene's spatial information; and (3) \textit{Temporal dependency}, involving streaming queries that rely on preceding context.}
    \label{fig:data_stat}
    \vspace{-3mm}
\end{figure}

\vspace{0.05in}\noindent\textbf{Active exploration.} The simulated part of $S^3$-Bench natively supports active exploration. To train this capability, we synthesize tailored trajectories and paired QA annotations. When keypoint sweeping is incomplete or target objects suffer from heavy occlusion, QA pairs are instantiated at those specific timestamps. The pipeline then generates subsequent camera movements (translation or rotation) driven by the need to capture additional information. These newly acquired observations are integrated into the training set, with the corresponding movements but not the answer serving as ground-truth action labels. During testing, the model's output actions are executed directly in the simulator to gather new visual context for question answering.

\subsection{Real-World Data Integration.}
To accurately assess the model's generalization capabilities in physical environments, $S^3$-Bench incorporates an extensive real-world subset. Real-world data align more closely with practical application scenarios and incorporate specific real-world artifacts, such as irregular camera motion and dynamic blur.  We aggregate continuous video streams and corresponding 3D scene metadata from established large-scale datasets, including ScanNet \cite{dai2017scannet}, ScanNet++ \cite{yeshwanth2023scannet++}, and ARKitScenes \cite{baruch2021arkitscenes}. This yields a collection of about 3,000 unique physical scenes and 6,000 fixed scan trajectories. Due to the inherent constraints of pre-recorded videos, the real-world domain is utilized to evaluate streaming spatial reasoning under fixed camera trajectories, complementing the active exploration evaluation conducted in the simulator. We leverage 3D scene annotations to derive object bounding boxes and frame-level object visibility. To implement streaming annotation, the generation process is constrained to using only visibility-related information available prior to the timestamp of the question.

\begin{table}[t]
    \centering
    \scriptsize 
    \caption{\small \textbf{Categories and examples of $S^3$-Bench QA pairs.} Tasks span three domains: Ego-centric, Spatial Relation, and Temporal Dependency. Each entry specifies the task, answer format (Numerical/Multiple Choice), and a question template.}
    \label{tab:qa_categories}
    \renewcommand{\arraystretch}{0.95} 
    \begin{tabularx}{\textwidth}{@{}lX@{}}
        \toprule
        \textbf{Task [Format]} & \textbf{Question} \\ 
        \midrule
        
        \multicolumn{2}{@{}l}{\textit{\textbf{Ego-centric}}} \\
        camera\_displacement [Num] & How far did the camera move from the past X seconds? \\
        current\_room\_area [Num] & What is the total area of the current room? \\
        cam\_obj\_distance [Num] & What is the distance between the camera and the {A}?\\
        identification\_closest [MC] & Which object is currently spatially closest to the camera? \\
        camera\_motion\_target [MC] & Which object did the camera approach the most? \\ 
        \midrule
        
        \multicolumn{2}{@{}l}{\textit{\textbf{Spatial Relation Understanding}}} \\
        attribute [Num] & What is the length of \{A\}'s longest edge? \\
        spatial\_distance [Num] & What is the distance between the \{A\} and the \{B\}? \\
        area [Num] & What is the total area of the rooms visited so far? \\
        count [Num] & How many \{A\}s are visible so far? \\
        sequence [MC] & Identify the correct chronological order of \{A, B, C\}. \\
        spatial\_proximity [MC] & Which object category is spatially closest to the \{A\}? \\
        relative\_orientation [MC] & In which direction is the \{C\} located? \\
        sequence\_identification [MC] & Which of the following objects appeared first? \\ 
        \midrule
        
        \multicolumn{2}{@{}l}{\textit{\textbf{Temporal Dependency}}} \\
        tem\_spatial\_distance\_ref [Num] & What is the distance from that object to the \{A\}? \\
        tem\_cam\_obj\_distance\_ref [Num] & How far (in meters) is the camera from it now? \\
        tem\_horizontal\_direction [MC] & Is that object located to the Front-Left, Front-Right, Back-Left or Back-Right? \\ 
        \bottomrule
    \end{tabularx}
    \vspace{-3mm}
\end{table}

\subsection{Streaming QA Annotation Pipeline}
To generate QA pairs that inherently require continuous temporal and spatial reasoning, we develop an automated, metadata-driven annotation pipeline that is applicable to both simulated and real-world domains. The process begins by extracting videos and meta-information (3D bounding boxes and frame-level object visibility) from the two data sources. We then sample various points in time to serve as question timestamps. QA pairs are generated according to templates and construction rules, strictly constrained to using visibility data from before each timestamp to maintain temporal consistency. Finally, to ensure linguistic diversity and mitigate template-induced biases, an advanced Large Language Model (e.g., Gemini \cite{team2023gemini}) is employed to refine the generated QA pairs, ensuring natural and varied linguistic expressions. Fig.\ref{fig:data_stat} illustrates the distribution and samples of QA pairs.

\subsection{Task Categorization and Evaluation Protocol}
To comprehensively evaluate the spatial and temporal reasoning capabilities of the models under continuous visual streams, the question-answer pairs in our benchmark are systematically categorized into three core dimensions. As detailed in Tab.\ref{tab:qa_categories}, the evaluation encompasses two distinct response formats to assess different levels of reasoning precision: Numerical tasks, which require quantitative output (e.g. exact distances, object counts, or spatial areas), and Multiple Choice tasks, which assess discrete spatial, topological, or temporal judgments. Tab.\ref{tab:qa_categories} shows all QA categories and templates.

\vspace{0.05in}\noindent\textit{Ego-centric Perception.} 
This dimension evaluates the dynamic spatial relationship between the embodied agent (ego) and its immediate surroundings. In strict alignment with the streaming setting, these tasks focus on processing real-time information located around the specific timestamp when the query is posed. Rather than requiring extensive historical recall, the model must assess its current state and local environment. Representative tasks involve calculating camera displacement over a recent temporal window, measuring the current room area, and identifying objects in immediate proximity to the camera. This category serves to validate the model's instantaneous spatial awareness and its capacity to process concurrent visual inputs.

\vspace{0.05in}\noindent\textit{Spatial Relation.} 
This category focuses on the structural and geometric comprehension of the scene itself. Distinct from the instantaneous nature of ego-centric perception, this dimension evaluates the model's ability to process and consolidate environmental information that appeared prior to the moment the question was asked. The model is required to construct a coherent spatial representation based on the historical context of the video stream. The core capabilities tested here include calculating physical distances between previously observed objects, determining the chronological sequence of object appearances, computing accumulated areas of visited spaces, and inferring the relative orientations of distinct targets. These tasks require reliable spatial memory and the ability to integrate past visual observations.

\vspace{0.05in}\noindent\textit{Temporal Dependency.} 
This dimension specifically assesses the model's capability to maintain sustained reasoning across multiple interactions, which is essential for multi-turn question-answering in streaming scenarios. In these tasks, the premise of the current question is intrinsically dependent on information established or entities identified in preceding queries. To answer accurately, the model must continuously correlate contextual information across the temporal sequence. For instance, the model may need to calculate the updated spatial distance or track the dynamic horizontal direction of a specific object referenced in a prior interaction. This category ensures the evaluation accurately reflects the model's capacity for contextual continuity and progressive temporal logic.

\vspace{0.05in}\noindent\textbf{Evaluation Protocol.} Following VSI-Bench \cite{yang2025thinking}, for multiple-choice questions, we employ standard Accuracy. For open-ended numerical estimations (e.g. distances, areas), we utilize Mean Relative Accuracy (MRA) to account for absolute scale variances. Crucially, for the Multi-Step Dependency Chains, we enforce a \textit{strict joint evaluation protocol}. During inference, Q2 is conditioned strictly on the model's own generative output from Q1, rather than being provided with ground-truth teacher forcing. This rigorous setup explicitly evaluates the model's vulnerability to error propagation and its capacity for coherent, long-horizon spatial reasoning. 
\section{AMF-VLM}

\begin{figure}[t]
    \centering
    \includegraphics[width=0.85\linewidth]{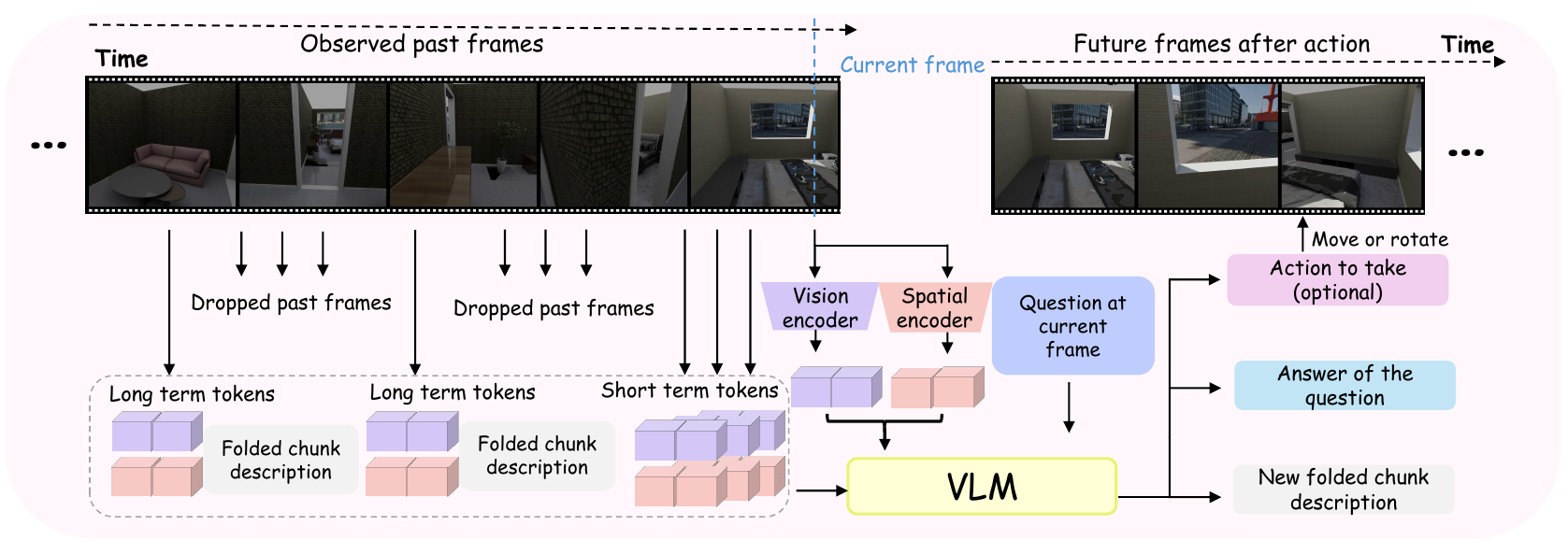}
    \vspace{-1.5mm}
    \caption{\small \textbf{AMF-VLM architecture.} Visual features are extracted via a vision encoder, optionally supplemented by a spatial encoder. For efficient streaming, recent frames are densely sampled while long-term history is sparsely retained and compressed via memory folding. Finally, the model outputs either an exploration action, a direct answer, or a memory update summary.}
    \label{fig:method}
    \vspace{-3mm}
\end{figure}

For streaming spatial understanding and active exploration, we propose the Active Memory-Folding Vision-Language Model (AMF-VLM). As illustrated in Fig.\ref{fig:method}, AMF-VLM efficiently processes continuous visual streams with a stable computational footprint. The framework integrates three core components: a streaming processing and training strategy; periodic memory folding for long-term context management; and an active exploration mechanism to acquire scene information when current observations are insufficient.

\subsection{Streaming Processing and Training Strategy}

AMF-VLM processes continuous video streams incrementally. To maintain a bounded Key-Value (KV) cache, we employ a hierarchical streaming mechanism incorporating attention sinks and continuous positional encoding \cite{xu2025streamingvlm}. Visual frames are managed by temporal proximity: recent windows retain high sampling density, while distant history is sub-sampled.

To stabilize long-sequence generation, fixed initial tokens are retained as attention sinks to prevent attention collapse. Continuous Rotary Position Embedding (RoPE) ensures positional continuity across shifting windows and prevents ID overflow \cite{xu2025streamingvlm}. To compensate for reduced visual density in distant frames, the model maintains corresponding textual descriptions as explicit text memory.

During training, long sequences are truncated into shorter overlapping samples to simulate streaming within memory constraints. This strategy aligns training with inference, enabling the model to learn localized attention patterns that are generalized to continuous streams.

\subsection{Periodic Memory Folding}
Explicitly reasoning over extended horizons requires maintaining a structured semantic understanding of the sub-sampled distant frames. We employ a Streaming Memory Folding mechanism, triggered periodically every $K$ seconds (empirically set to $K=10$). After each folding operation, the system retains only a single visual frame along with its corresponding textual description. For short-term context, information is typically preserved at a sampling rate of 1 FPS within a 10-second window.

During a folding step at time $t$, the VLM generates a local textual description $D_t$ of the scene observed within the recent window. This description is recursively merged with the previous global memory state $M_{t-K}$ to synthesize an updated global scene summary. 

To provide stable historical context and mitigate information loss over long trajectories, the generated text memory is integrated into the visual sequence. Specifically, at inference step $t$, the textual memory summaries $\mathcal{M}$ are interleaved chronologically within the sub-sampled distant frames $\mathcal{V}_{sparse}$. This interleaved formulation explicitly aligns the sparse visual signals with their corresponding semantic anchors. The final input sequence $S_t$ is defined as:
\begin{equation*}
S_t = [\text{Prompt}, \text{Interleave}(\mathcal{V}_\text{sparse}, \mathcal{M}), \mathcal{V}_\text{dense}, Q_t]
\end{equation*}

where prompt is the system prompt, $\mathcal{V}_\text{dense}$ denotes the high-density recent frames and $Q_t$ represents the current query.

To train this memory summarization capability, we construct a dedicated dataset of approximately 50,000 samples. We utilize short video segments and automatically generated concise ground-truth scene descriptions based on scene meta-information utilizing the Gemini model.

\subsection{Active Exploration Mechanism}
Current video trajectories may not capture sufficient visual information. AMF-VLM integrates an active exploration mechanism to autonomously acquire supplementary viewpoints and then give the final answer. The egocentric action space $\mathcal{A}$ comprises parameterized translational movements $\mathcal{A}_\text{trans}$ and discrete rotational adjustments $\mathcal{A}_\text{rot}$. For translational movements, each action specifies a displacement distance $d$ (e.g., move forward $d$, move left $d$, move right $d$). Rotational adjustments include rotate left $45^\circ$, rotate right $45^\circ$, scan forward ($90^\circ$), and sweep around ($360^\circ$). 

When the model determines that the current observation and memory are insufficient to answer a query, it outputs an exploration command $a_t \in \mathcal{A}$. To ensure standardization, the action output, including the specific movement direction and the distance parameter $d$, is structured in JSON format.

To supervise this active behavior, we generated approximately 100,000 training samples using simulator. These samples were systematically constructed using predefined rules to create scenarios with insufficient observational information (e.g., target objects out of view or occluded). Each sample is annotated with the corresponding corrective action in JSON format, teaching the model to actively explore rather than output hallucinated responses.

\subsection{Optional 3D Spatial Integration}
While AMF-VLM primarily relies on 2D visual streams and memory folding, it supports the optional integration of explicit 3D spatial features. If external 3D information is available, incorporating spatial features $f_\text{spatial}^{(t)}$ and camera pose features $f_\text{cam}^{(t)}$ can enhance the model's 3D spatial understanding capabilities to a certain extent. These features are processed via a StreamVGGT \cite{zhuo2025streaming} module and projected into the latent space of LLM to generate spatial tokens, formulated as $\text{token}_\text{spatial}^{(t)} = \text{MLP}(f_\text{spatial}^{(t)} \oplus f_\text{cam}^{(t)})$, where $\oplus$ denotes concatenation. These spatial tokens can then be appended to the visual input sequence to provide supplementary geometric context.
\begin{table*}[t]
\centering
\caption{\small Performance comparison of baselines across all $S^3$-Eval simulation part. (*) indicates models also trained on our data. ($\dagger$) indicates that the model is modified as streaming inference.}
\vspace{-1.5mm}
\label{tab:performance_sim}
\tiny
\setlength{\tabcolsep}{1.35pt} 
\renewcommand{\arraystretch}{1.1} 

\begin{tabular}{l | *{17}{c}}
\toprule
Method & Over- & T-C & T- & T-Spa & Spa- & Spa- & C- & C- & C- & Id- & Rel- & Area & C- & Attr & Cou & Seq & Seq \\
 & all & -O-D & Hor & -D & D & Pro & O-D & Dist & Mo & Clo & Ori &  & Area &  &  &  & Id \\
\midrule
Random & 25.6 & - & 21.0 & - & - & 29.1 & - & - & 23.7 & 22.2 & 30.7 & - & - & - & - & 26.2 & 26.6 \\
Frequent & 35.6 & 46.1 & 26.1 & 47.6 & 46.7 & 26.0 & 33.0 & 39.3 & 26.9 & 27.7 & 37.1 & 37.6 & 35.9 & 46.5 & 39.6 & 26.8 & 25.8 \\
Human level & 80.5 & 71.1 & 92.6 & 70.4 & 70.9 & 93.0 & 71.3 & 71.7 & 98.6 & 92.6 & 97.1 & 73.4 & 70.9 & 66.1 & 66.0 & 93.5 & 92.8 \\
\midrule
GPT-5.2 \cite{achiam2023gpt} & 44.9 & 38.4 & 29.7 & 37.4 & 34.6 & 41.8 & 35.9 & 45.5 & 61.4 & 58.6 & 44.4 & 35.6 & 44.4 & 51.5 & 53.4 & 49.6 & 52.7 \\
Gemini-3-flash \cite{team2023gemini} & 41.9 & 35.6 & 27.6 & 34.5 & 31.5 & 38.8 & 32.7 & 42.3 & 58.1 & 55.5 & 41.5 & 32.5 & 41.6 & 48.5 & 49.6 & 46.8 & 49.6 \\
Gemini-3-pro \cite{team2023gemini} & 43.1 & 27.1 & 32.5 & 47.6 & 24.2 & 58.3 & 24.9 & 32.2 & 69.1 & 59.0 & 31.4 & 53.7 & 58.0 & 36.7 & 37.8 & 55.5 & 37.6 \\
\midrule
llava-video \cite{zhang2024llava} & 31.6 & 7.03 & 22.8 & 11.4 & 29.7 & 38.4 & 21.3 & 15.5 & 50.6 & 45.5 & 46.3 & 14.0 & 29.2 & 39.2 & 52.6 & 35.5 & 47.1 \\
llava-ov-1.5-4B \cite{an2025llava}& 22.4 & 5.53 & 27.4 & 26.5 & 16.6 & 38.1 & 2.71 & 2.22 & 31.6 & 36.6 & 39.8 & 7.32 & 30.4 & 0.61 & 29.0 & 29.0 & 37.1 \\
llava-ov-1.5-8B \cite{an2025llava} & 28.8 & 7.89 & 24.6 & 16.2 & 33.8 & 38.6 & 26.1 & 5.73 & 32.3 & 41.2 & 41.7 & 12.1 & 31.9 & 42.0 & 31.0 & 34.4 & 44.0 \\

Qwen3-VL-4B \cite{bai2025qwen3} & 30.8 & 27.5 & 20.7 & 35.6 & 30.2 & 26.5 & 27.0 & 41.0 & 14.8 & 19.2 & 26.5 & 31.4 & 36.5 & 43.7 & 42.1 & 38.9 & 31.7 \\
Qwen3-VL-8B \cite{bai2025qwen3} & 41.1 & 36.0 & 25.8 & 34.0 & 29.7 & 39.6 & 32.0 & 41.1 & 56.7 & 54.2 & 40.2 & 32.5 & 41.5 & 48.1 & 48.8 & 45.4 & 49.1 \\
Qwen3-VL-32B \cite{bai2025qwen3} & 37.4 & 34.0 & 27.1 & 37.8 & 31.9 & 37.9 & 38.5 & 40.0 & 52.2 & 48.8 & 50.1 & 46.8 & 43.4 & 1.53 & 42.9 & 13.2 & 40.2 \\

InternVL3.5-1B \cite{wang2025internvl3} & 26.4 & 17.7 & 26.1 & 12.9 & 11.8 & 35.9 & 10.5 & 22.2 & 49.2 & 43.1 & 43.0 & 0.94 & 20.3 & 0.22 & 52.4 & 31.8 & 44.3 \\
InternVL3.5-8B \cite{wang2025internvl3} & 37.7 & 23.3 & 25.6 & 39.9 & 27.7 & 42.0 & 31.9 & 34.0 & 49.9 & 45.8 & 46.7 & 28.9 & 42.9 & 13.8 & 56.4 & 43.9 & 48.9 \\
InternVL3.5-38B \cite{wang2025internvl3} & 35.2 & 43.4 & 27.1 & 28.4 & 27.2 & 41.5 & 21.1 & 37.4 & 37.9 & 44.2 & 52.0 & 23.3 & 36.3 & 0.16 & 52.7 & 39.4 & 49.4 \\
\midrule
vlm-3r \cite{fan2025vlm} & 41.9 & 31.6 & 27.1 & 35.2 & 32.3 & 40.8 & 32.5 & 41.1 & 56.2 & 53.6 & 41.1 & 41.4 & 43.0 & 49.2 & 49.1 & 44.5 & 47.3 \\
cambrian-s \cite{yang2025cambrian} & 44.8 & 29.2 & 22.8 & 48.1 & 50.1 & 37.9 & 30.3 & 58.7 & 50.4 & 52.5 & 37.3 & 58.0 & 38.8 & 43.4 & 58.1 & 39.2 & 56.3 \\
\midrule
cambrian-s*$^{\dagger}$ \cite{yang2025cambrian} & 57.8 & 52.1 & 33.5 & 51.3 & 53.2 & 62.4 & 59.8 & 62.1 & 70.5 & 67.8 & 45.9 & 58.4 & 61.3 & 64.3 & 59.2 & 58.3 & 58.8 \\
streaming-vlm* \cite{xu2025streamingvlm} & 56.0 & 54.9 & 31.0 & 51.8 & 45.2 & 58.3 & 66.0 & 60.9 & 67.0 & 65.6 & 40.0 & 59.1 & 60.1 & 63.2 & 55.8 & 55.2 & 55.0 \\ \\
\midrule
Ours & \textbf{62.9} & \textbf{57.6} & \textbf{35.0} & \textbf{55.6} & \textbf{59.4} & \textbf{70.4} & \textbf{70.8} & \textbf{67.7} & \textbf{75.6} & \textbf{72.8} & \textbf{46.3} & \textbf{67.0} & \textbf{68.1} & \textbf{70.8} & \textbf{61.8} & \textbf{60.9} & \textbf{60.6} \\
\bottomrule
\end{tabular}
\vspace{-3mm}
\end{table*}

\begin{table*}[t]
\centering
\caption{\small Performance comparison of baselines across all $S^3$-Eval real part. (*) indicates models also trained on our data. ($\dagger$) indicates streaming inference modification.}
\vspace{-1.5mm}
\label{tab:performance_real}
\tiny
\setlength{\tabcolsep}{1.5pt} 
\renewcommand{\arraystretch}{1.1} 

\begin{tabular}{l | *{16}{c}}
\toprule
Method & Over- & T-C & T- & T-Spa & Spa- & Spa- & C- & C- & C- & Id- & Rel- & Area  & Attr & Cou & Seq & Seq \\
 & all & -O-D & Hor & -D & D & Pro & O-D & Dist & Mo & Clo & Ori &   &  &  &  & Id \\
\midrule
Random & 26.1 & - & 23.1 & - & - & 25.2 & - & - & 25.1 & 24.6 & 34.4 & - & - & - & 23.5 & 24.5 \\
Frequent & 29.0 & 25.3 & 25.8 & 32.6 & 34.0 & 25.9 & 33.5 & 24.1 & 25.9 & 26.1 & 33.7 & 11.0 & 29.5 & 37.5 & 26.8 & 26.9 \\
Human level & 82.1 & 69.1 & 90.8 & 69.1 & 74.7 & 91.5 & 68.5 & 77.5 & 91.9 & 91.3 & 95.6 & 75.8 & 69.1 & 74.3 & 96.1 & 91.9 \\
\midrule
GPT-5.2 \cite{achiam2023gpt} & 41.7 & 28.1 & 29.8 & 31.5 & 25.8 & 59.3 & 25.5 & 39.5 & 61.8 & 51.4 & 44.7 & 48.2 & 27.8 & 54.7 & 49.5 & 52.2 \\
Gemini-3-flash \cite{team2023gemini} & 38.2 & 24.6 & 27.2 & 26.7 & 22.9 & 56.3 & 21.9 & 36.2 & 57.2 & 48.2 & 41.6 & 47.2 & 23.1 & 51.4 & 46.6 & 48.9 \\
Gemini-3-pro \cite{team2023gemini} & 40.4 & 20.8 & 22.7 & 22.9 & 39.9 & 45.2 & 36.7 & 29.8 & 69.5 & 46.7 & 44.1 & 49.5 & 53.0 & 41.5 & 37.4 & 59.1 \\
\midrule
llava-video \cite{zhang2024llava} & 34.9 & 14.8 & 33.1 & 11.5 & 21.2 & 52.3 & 18.6 & 32.6 & 63.4 & 46.9 & 44.0 & 18.2 & 17.9 & 51.3 & 38.4 & 49.3 \\
llava-ov-1.5-4B \cite{an2025llava} & 30.1 & 1.7 & 29.1 & 23.8 & 11.8 & 44.0 & 0.6 & 61.4 & 45.5 & 39.6 & 44.5 & 22.1 & 3.8 & 42.1 & 24.8 & 41.2 \\
llava-ov-1.5-8B \cite{an2025llava}& 34.2 & 17.3 & 27.1 & 21.1 & 31.1 & 49.6 & 24.3 & 38.4 & 53.7 & 44.3 & 41.2 & 19.6 & 18.3 & 36.6 & 32.8 & 44.6 \\

Qwen3-VL-4B \cite{bai2025qwen3}& 29.9 & 23.9 & 21.9 & 29.0 & 28.2 & 39.5 & 18.1 & 30.7 & 31.7 & 25.3 & 28.4 & 37.7 & 25.0 & 49.5 & 46.0 & 26.6 \\
Qwen3-VL-8B \cite{bai2025qwen3} & 39.0 & 25.3 & 27.7 & 27.6 & 23.7 & 57.3 & 22.1 & 36.1 & 59.8 & 49.3 & 41.6 & 48.5 & 25.0 & 51.9 & 47.3 & 50.0 \\
Qwen3-VL-32B \cite{bai2025qwen3}& 37.3 & 34.1 & 25.2 & 38.1 & 28.2 & 53.0 & 35.6 & 29.6 & 57.5 & 42.9 & 54.1 & 46.2 & 21.9 & 48.3 & 6.3 & 28.5 \\

InternVL3.5VL-1B \cite{wang2025internvl3} & 36.2 & 20.5 & 36.6 & 17.8 & 21.8 & 51.2 & 13.6 & 39.1 & 62.3 & 45.5 & 44.2 & 33.7 & 20.9 & 52.2 & 36.1 & 47.0 \\
InternVL3.5VL-8B \cite{wang2025internvl3} & 40.5 & 19.7 & 27.8 & 37.5 & 35.0 & 53.8 & 26.6 & 38.6 & 58.0 & 47.6 & 44.3 & 43.9 & 24.6 & 55.2 & 42.5 & 50.4 \\
InternVL3.5VL-32B \cite{wang2025internvl3} & 41.3 & 35.8 & 29.1 & 30.6 & 32.5 & 56.8 & 32.7 & 19.4 & 58.6 & 49.2 & 53.2 & 42.9 & 22.8 & 55.0 & 50.1 & 50.6 \\
\midrule
vlm-3r \cite{fan2025vlm} & 42.1 & 34.1 & 28.5 & 30.0 & 33.9 & 53.0 & 30.4 & 39.0 & 59.2 & 48.1 & 44.4 & 41.0 & 42.8 & 52.5 & 51.0 & 45.8 \\
cambrian-s \cite{yang2025cambrian} & 45.1 & 39.8 & 34.7 & 25.3 & 45.0 & 65.9 & 40.4 & 32.3 & 69.9 & 46.7 & 44.4 & 34.9 & 60.8 & 43.2 & 64.8 & 37.2 \\
\midrule
cambrian-s*$^{\dagger}$ \cite{yang2025cambrian} & 51.0 & 38.4 & 33.0 & 41.9 & 44.7 & 65.1 & 38.8 & 55.2 & 70.9 & 54.8 & 48.2 & 45.4 & 58.2 & 59.3 & 56.2 & 54.8 \\
streaming-vlm* \cite{xu2025streamingvlm} & 49.1 & 35.2 & 30.4 & 38.6 & 41.3 & 62.3 & 35.5 & 51.9 & 67.4 & 51.9 & 46.1 & 42.9 & 54.4 & 61.8 & 64.8 & 52.6 \\
\midrule
Ours & \textbf{57.8} & \textbf{41.0} & \textbf{37.0} & \textbf{48.2} & \textbf{50.1} & \textbf{71.8} & \textbf{45.1} & \textbf{77.5} & \textbf{75.5} & \textbf{59.7} & \textbf{44.5} & \textbf{49.7} & \textbf{70.9} & \textbf{64.0} & \textbf{72.0} & \textbf{64.4} \\
\bottomrule
\end{tabular}
\vspace{-3mm}
\end{table*}

\begin{table}[t]
    \centering
    \begin{minipage}[t]{0.48\textwidth}
        \centering
        \scriptsize
        \caption{\small \textbf{Component ablation.} Active Exploring and Memory Folding improve accuracy in specific scenarios, while the optional 3D encoder enhances spatial understanding to a certain extent.}
        \label{tab:ablation}
        \vspace{-2mm}
        \begin{tabular*}{\linewidth}{@{\extracolsep{\fill}}lccc}
            \toprule
            Method & Numerical &Multi-choice & Overall \\
            \midrule
            Baseline (data)  & 57.4 & 53.1 & 56.0 \\
            \quad + active   & 60.1 & 54.6 & 58.1 \\
            \quad + folding  & 62.5 & 58.0 & 60.9 \\
            \quad + 3D enc   & 64.3 & 60.2 & 62.9 \\
            \bottomrule
        \end{tabular*}

        \caption{\small \textbf{Latency analysis(Time in s).} Unlike standard non-streaming models that scale poorly with input length, our method maintains low, stable latency across increasing frame counts.}
        \label{tab:latency}
        \vspace{-3mm}
        \begin{tabular*}{\linewidth}{@{\extracolsep{\fill}}lccccc}
            \toprule
            
            Method / Frames & 8 & 16 & 32 & 64 & 128 \\
            \midrule
            Qwen3-VL-8B & 2.03 & 2.19 & 2.70 & 3.22 & 8.56 \\
            Ours-8B     & 0.42 & 0.43 & 0.48 & 0.51 & 0.53 \\
            \bottomrule
        \end{tabular*}
    \end{minipage}\hfill
    \begin{minipage}[t]{0.48\textwidth}
    
        \centering
        \small
        \caption{\small \textbf{Results on standard spatial benchmarks.} Our method achieves competitive offline performance. The baseline prior to training on our dataset (*) highlights the consistent transfer gains provided by our constructed data.}
        \label{tab:other_bench}
        \vspace{2pt}
        \renewcommand{\arraystretch}{1.45} 
        \begin{tabular*}{\linewidth}{@{\extracolsep{\fill}}lccc}
            \toprule
            Method & VSIbench & Blink & MMSI \\
            \midrule
            GPT5            & 55.0 & 62.8 & 41.8 \\
            Gemini-2.5-Pro  & 53.5 & 70.0 & 38.0 \\
            Qwen3-VL-8b     & 59.4 & 69.1 & 31.1 \\
            InternVL-3.5-8b & 56.3 & 59.5 & 27.3 \\
            ours-w.o.-data* & 55.0 & 62.4 & 30.5 \\
            ours            & 67.5 & 65.1 & 31.2 \\
            \bottomrule
        \end{tabular*}
    \end{minipage}
\end{table}

\section{Experiments}
\label{sec:experiments}

\subsection{Experimental Setup}
\textbf{Implementation Details.} The training data are a combination of our constructed streaming spatial data, active exploration data, memory folding data, and VSI590K \cite{yang2025cambrian} dataset. We adopt Qwen3-VL-8B \cite{bai2025qwen3} as the base model. Parameters of the vision encoder are frozen. The model is trained for 1 epoch with a batch size of 16. More details and experiments are shown in the Appendix.

\subsection{Baselines}
We compare against five baseline categories: Heuristic and Human (Random Guess, Frequent Answer, 10\% subset for human); Closed-source (GPT-5.2 \cite{achiam2023gpt}, Gemini 3 Flash/Pro \cite{team2023gemini}); Open-source (LLaVA-OV-1.5 \cite{an2025llava}, LLaVA-Video \cite{zhang2024llava}, Qwen3-VL \cite{bai2025qwen3}, InternVL3.5 \cite{wang2025internvl3}); spatial model (VLM-3R \cite{fan2025vlm}, Cambrian-S \cite{yang2025cambrian}); and streaming model(Streaming-VLM \cite{xu2025streamingvlm}, Cambrian-S with modified streaming inference). The streaming models are also trained with our data. We also evaluate the streaming models after training on our dataset.

For non-streaming models, the visual input is restricted to a maximum of 32 frames sampled prior to question timestamp. For temporal dependency questions, preceding question-answer pairs are provided in the prompt as context.

\subsection{Main Results}
\noindent\textbf{Performance on $S^3$-eval.} We evaluate the models on our streaming spatial understanding benchmark, the $S^3$-Eval, which consists of two distinct splits: Sim (simulated environments) as shown in Tab.\ref{tab:performance_sim} and Real (real-world scenarios) as shown in Tab.\ref{tab:performance_real}. Our proposed method demonstrates strong performance across both splits. Furthermore, other streaming models exhibit performance improvements when trained on our dataset. The results also indicate that spatial models outperform standard open-source models that lack specific spatial designs, confirming the necessity of spatial awareness designs.

\noindent\textbf{Performance on Other Spatial Benchmarks.} On common spatial benchmarks, including VSI-Bench \cite{yang2025thinking}, Blink \cite{fu2024blink}, and MMSI\cite{yang2025mmsi}, as shown in Tab.\ref{tab:other_bench}, our trained model achieves results comparable to established baselines (baseline performances are cited from other publications \cite{bai2025qwen3, wang2025internvl3, cai2025scaling}). This further verifies the effectiveness of our training data and approach.

\subsection{Ablation Study}
As shown in Tab.\ref{tab:ablation}, we conduct ablation experiments to validate the contributions of the proposed modules. The Active Exploring mechanism improves the model's accuracy on questions requiring detailed environmental observation, such as identifying attributes of objects or determining the room area. Memory Folding effectively retains structural context, which specifically benefits tasks relying on long-term spatial information. Furthermore, if explicit 3D features are provided, the 3D Encoder can improve the 3D spatial understanding capabilities to a certain extent, serving as an effective optional component. More detailed ablation studies and quantitative results are provided in the Appendix.

\subsection{Efficiency Analysis}
We measure the answering latency of our method and compare it with Qwen3-VL-8B using a standard inference strategy. As shown in Tab\ref{tab:latency}, the standard Qwen3-VL-8B model cannot process the video stream in real-time, as its latency increases significantly with the number of input frames. In contrast, our method maintains consistently low latency, and the processing time increases slowly as the frame count grows, meeting the practical requirements for continuous online understanding.

\begin{figure}[t]
    \centering
    \includegraphics[width=0.95\linewidth]{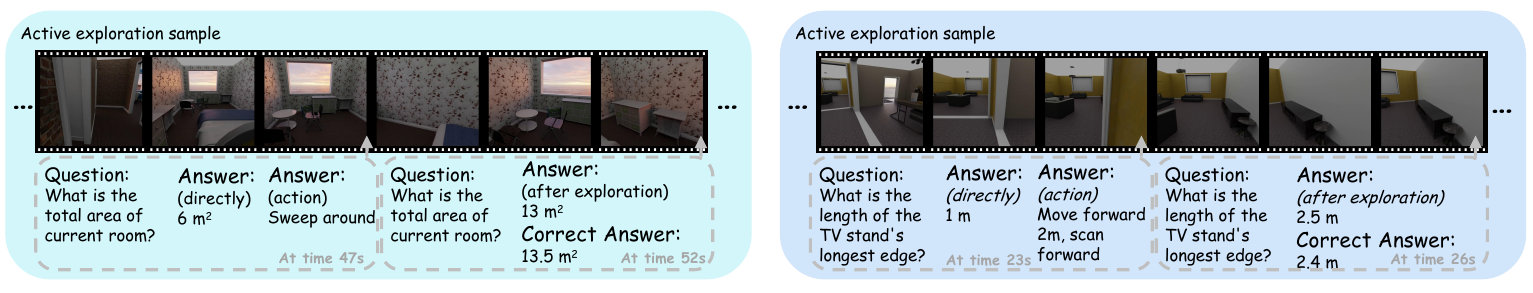}
    \vspace{-1.5mm}
    \caption{\small \textbf{Qualitative examples of Active Exploration.} Restricted initial views can lead to incorrect direct predictions. To mitigate this, the model executes exploration actions to gather sufficient visual context before responding.}
    \label{fig:active_sample}
    \vspace{-3mm}
\end{figure}

\begin{figure}[t]
    \centering
    \includegraphics[width=0.95\linewidth]{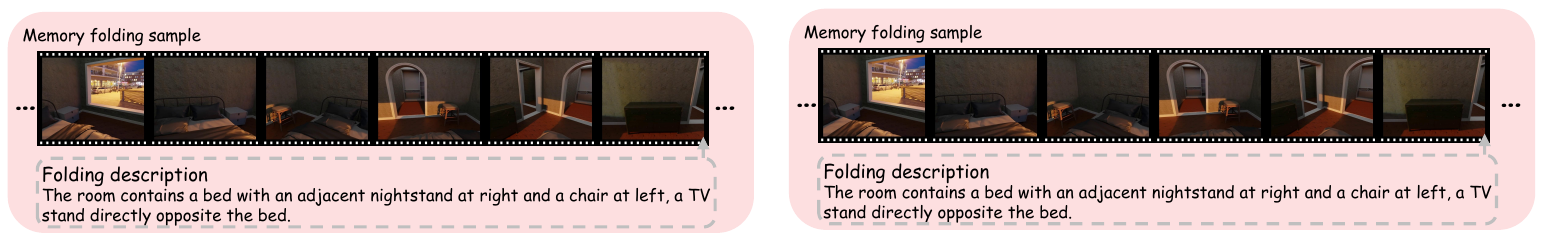}
    \vspace{-1.5mm}
    \caption{\small \textbf{Qualitative examples of Memory Folding.} AMF-VLM periodically compresses observed scenes into concise text, efficiently retaining long-term spatial memory and mitigating forgetting over extended sequences.}
    \label{fig:memory_sample}
    \vspace{-3mm}
\end{figure}

\subsection{Qualitative Analysis}
\textbf{Active Exploring.}  We visualize the effect of the Active Exploration module as shown in Fig.\ref{fig:active_sample}. Without this mechanism, the model lacks sufficient observational vision input and fails to predict the correct answer. Active Exploration enables the model to gather the necessary spatial and environmental information to answer the question correctly.

\noindent\textbf{Memory Folding.}  We provide examples of the Memory Folding process as shown in Fig.\ref{fig:memory_sample}. This module generates concise descriptions of the observed scene, which helps the model retain structural information and reduces long-term forgetting during extended video inputs.
\section{Conclusion}

In this work, we deviate from conventional offline spatial understanding by formulating \textit{streaming spatial understanding} based on real-world application situations. We introduce $S^3$-Bench, a comprehensive benchmark encompassing diverse simulated and real-world environments. $S^3$-Bench features streaming QA pairs and active exploration, providing both training and evaluation subsets. Furthermore, we propose AMF-VLM, a model designed for streaming spatial reasoning and active exploration. By incorporating a \textit{memory folding} mechanism, AMF-VLM achieves efficient long-horizon memory. When the observation is insufficient to respond to the query, AMF-VLM outputs actions to actively acquire additional spatial information. Experimental results demonstrate that AMF-VLM achieves strong performance on $S^3$-Bench and remains competitive on standard spatial understanding benchmarks. In the future we will explore the integration of streaming spatial understanding with broader streaming-based embodied intelligence tasks.

\bibliographystyle{splncs04}
\bibliography{main}

@inproceedings{majumdar2024openeqa,
  title={Openeqa: Embodied question answering in the era of foundation models},
  author={Majumdar, Arjun and Ajay, Anurag and Zhang, Xiaohan and Putta, Pranav and Yenamandra, Sriram and Henaff, Mikael and Silwal, Sneha and Mcvay, Paul and Maksymets, Oleksandr and Arnaud, Sergio and others},
  booktitle={Proceedings of the IEEE/CVF conference on computer vision and pattern recognition},
  pages={16488--16498},
  year={2024}
}

@inproceedings{azuma2022scanqa,
  title={Scanqa: 3d question answering for spatial scene understanding},
  author={Azuma, Daichi and Miyanishi, Taiki and Kurita, Shuhei and Kawanabe, Motoaki},
  booktitle={proceedings of the IEEE/CVF conference on computer vision and pattern recognition},
  pages={19129--19139},
  year={2022}
}

@article{cheng2024spatialrgpt,
  title={Spatialrgpt: Grounded spatial reasoning in vision-language models},
  author={Cheng, An-Chieh and Yin, Hongxu and Fu, Yang and Guo, Qiushan and Yang, Ruihan and Kautz, Jan and Wang, Xiaolong and Liu, Sifei},
  journal={Advances in Neural Information Processing Systems},
  volume={37},
  pages={135062--135093},
  year={2024}
}

@article{liu2023visual,
  title={Visual spatial reasoning},
  author={Liu, Fangyu and Emerson, Guy and Collier, Nigel},
  journal={Transactions of the Association for Computational Linguistics},
  volume={11},
  pages={635--651},
  year={2023},
  publisher={MIT Press One Broadway, 12th Floor, Cambridge, Massachusetts 02142, USA~…}
}

@inproceedings{ma20253dsrbench,
  title={3dsrbench: A comprehensive 3d spatial reasoning benchmark},
  author={Ma, Wufei and Chen, Haoyu and Zhang, Guofeng and Chou, Yu-Cheng and Chen, Jieneng and de Melo, Celso and Yuille, Alan},
  booktitle={Proceedings of the IEEE/CVF International Conference on Computer Vision},
  pages={6924--6934},
  year={2025}
}

@article{xu2024vlm,
  title={Vlm-grounder: A vlm agent for zero-shot 3d visual grounding},
  author={Xu, Runsen and Huang, Zhiwei and Wang, Tai and Chen, Yilun and Pang, Jiangmiao and Lin, Dahua},
  journal={arXiv preprint arXiv:2410.13860},
  year={2024}
}

@article{yang2025cambrian,
  title={Cambrian-s: Towards spatial supersensing in video},
  author={Yang, Shusheng and Yang, Jihan and Huang, Pinzhi and Brown, Ellis and Yang, Zihao and Yu, Yue and Tong, Shengbang and Zheng, Zihan and Xu, Yifan and Wang, Muhan and others},
  journal={arXiv preprint arXiv:2511.04670},
  year={2025}
}

@article{fan2025vlm,
  title={Vlm-3r: Vision-language models augmented with instruction-aligned 3d reconstruction},
  author={Fan, Zhiwen and Zhang, Jian and Li, Renjie and Zhang, Junge and Chen, Runjin and Hu, Hezhen and Wang, Kevin and Qu, Huaizhi and Wang, Dilin and Yan, Zhicheng and others},
  journal={arXiv preprint arXiv:2505.20279},
  year={2025}
}

@article{lin2025mmsi,
  title={MMSI-Video-Bench: A Holistic Benchmark for Video-Based Spatial Intelligence},
  author={Lin, Jingli and Xu, Runsen and Zhu, Shaohao and Yang, Sihan and Cao, Peizhou and Ran, Yunlong and Hu, Miao and Zhu, Chenming and Xie, Yiman and Long, Yilin and others},
  journal={arXiv preprint arXiv:2512.10863},
  year={2025}
}

@inproceedings{yang2025thinking,
  title={Thinking in space: How multimodal large language models see, remember, and recall spaces},
  author={Yang, Jihan and Yang, Shusheng and Gupta, Anjali W and Han, Rilyn and Fei-Fei, Li and Xie, Saining},
  booktitle={Proceedings of the Computer Vision and Pattern Recognition Conference},
  pages={10632--10643},
  year={2025}
}

@article{zhao2025vla,
  title={VLA-RAIL: A Real-Time Asynchronous Inference Linker for VLA Models and Robots},
  author={Zhao, Yongsheng and Zhao, Lei and Cheng, Baoping and Yao, Gongxin and Wen, Xuanzhang and Gao, Han},
  journal={arXiv preprint arXiv:2512.24673},
  year={2025}
}

@article{mavrogiannis2023core,
  title={Core challenges of social robot navigation: A survey},
  author={Mavrogiannis, Christoforos and Baldini, Francesca and Wang, Allan and Zhao, Dapeng and Trautman, Pete and Steinfeld, Aaron and Oh, Jean},
  journal={ACM Transactions on Human-Robot Interaction},
  volume={12},
  number={3},
  pages={1--39},
  year={2023},
  publisher={ACM New York, NY}
}

@article{zhao2025spacemind,
  title={SpaceMind: Camera-Guided Modality Fusion for Spatial Reasoning in Vision-Language Models},
  author={Zhao, Ruosen and Zhang, Zhikang and Xu, Jialei and Chang, Jiahao and Chen, Dong and Li, Lingyun and Sun, Weijian and Wei, Zizhuang},
  journal={arXiv preprint arXiv:2511.23075},
  year={2025}
}

@article{qi2025gpt4scene,
  title={Gpt4scene: Understand 3d scenes from videos with vision-language models},
  author={Qi, Zhangyang and Zhang, Zhixiong and Fang, Ye and Wang, Jiaqi and Zhao, Hengshuang},
  journal={arXiv preprint arXiv:2501.01428},
  year={2025}
}

@article{wu2025spatial,
  title={Spatial-mllm: Boosting mllm capabilities in visual-based spatial intelligence},
  author={Wu, Diankun and Liu, Fangfu and Hung, Yi-Hsin and Duan, Yueqi},
  journal={arXiv preprint arXiv:2505.23747},
  year={2025}
}

@inproceedings{fu2024blink,
  title={Blink: Multimodal large language models can see but not perceive},
  author={Fu, Xingyu and Hu, Yushi and Li, Bangzheng and Feng, Yu and Wang, Haoyu and Lin, Xudong and Roth, Dan and Smith, Noah A and Ma, Wei-Chiu and Krishna, Ranjay},
  booktitle={European Conference on Computer Vision},
  pages={148--166},
  year={2024},
  organization={Springer}
}

@article{yang2025mmsi,
  title={Mmsi-bench: A benchmark for multi-image spatial intelligence},
  author={Yang, Sihan and Xu, Runsen and Xie, Yiman and Yang, Sizhe and Li, Mo and Lin, Jingli and Zhu, Chenming and Chen, Xiaochen and Duan, Haodong and Yue, Xiangyu and others},
  journal={arXiv preprint arXiv:2505.23764},
  year={2025}
}

@inproceedings{chen2024spatialvlm,
  title={Spatialvlm: Endowing vision-language models with spatial reasoning capabilities},
  author={Chen, Boyuan and Xu, Zhuo and Kirmani, Sean and Ichter, Brain and Sadigh, Dorsa and Guibas, Leonidas and Xia, Fei},
  booktitle={Proceedings of the IEEE/CVF Conference on Computer Vision and Pattern Recognition},
  pages={14455--14465},
  year={2024}
}

@article{hu2025g,
  title={G$^2$ VLM: Geometry Grounded Vision Language Model with Unified 3D Reconstruction and Spatial Reasoning},
  author={Hu, Wenbo and Lin, Jingli and Long, Yilin and Ran, Yunlong and Jiang, Lihan and Wang, Yifan and Zhu, Chenming and Xu, Runsen and Wang, Tai and Pang, Jiangmiao},
  journal={arXiv preprint arXiv:2511.21688},
  year={2025}
}

@article{gholami2025spatial,
  title={Spatial reasoning with vision-language models in ego-centric multi-view scenes},
  author={Gholami, Mohsen and Rezaei, Ahmad and Weimin, Zhou and Mao, Sitong and Zhou, Shunbo and Zhang, Yong and Akbari, Mohammad},
  journal={arXiv preprint arXiv:2509.06266},
  year={2025}
}

@article{qu2025spatialvla,
  title={Spatialvla: Exploring spatial representations for visual-language-action model},
  author={Qu, Delin and Song, Haoming and Chen, Qizhi and Yao, Yuanqi and Ye, Xinyi and Ding, Yan and Wang, Zhigang and Gu, JiaYuan and Zhao, Bin and Wang, Dong and others},
  journal={arXiv preprint arXiv:2501.15830},
  year={2025}
}

@inproceedings{cai2025spatialbot,
  title={Spatialbot: Precise spatial understanding with vision language models},
  author={Cai, Wenxiao and Ponomarenko, Iaroslav and Yuan, Jianhao and Li, Xiaoqi and Yang, Wankou and Dong, Hao and Zhao, Bo},
  booktitle={2025 IEEE International Conference on Robotics and Automation (ICRA)},
  pages={9490--9498},
  year={2025},
  organization={IEEE}
}

@article{sun2025scaling,
  title={Scaling long-horizon llm agent via context-folding},
  author={Sun, Weiwei and Lu, Miao and Ling, Zhan and Liu, Kang and Yao, Xuesong and Yang, Yiming and Chen, Jiecao},
  journal={arXiv preprint arXiv:2510.11967},
  year={2025}
}

@article{su2026u,
  title={U-Fold: Dynamic Intent-Aware Context Folding for User-Centric Agents},
  author={Su, Jin and Fang, Runnan and Li, Yeqiu and Wang, Xiaobin and Cai, Shihao and Xie, Pengjun and Zhang, Ningyu and Yuan, Fajie},
  journal={arXiv preprint arXiv:2601.18285},
  year={2026}
}

@article{shao2025foldact,
  title={FoldAct: Efficient and Stable Context Folding for Long-Horizon Search Agents},
  author={Shao, Jiaqi and Miao, Yufeng and Zhang, Wei and Luo, Bing},
  journal={arXiv preprint arXiv:2512.22733},
  year={2025}
}

@inproceedings{chuang2025active,
  title={Active vision might be all you need: Exploring active vision in bimanual robotic manipulation},
  author={Chuang, Ian and Lee, Andrew and Gao, Dechen and Naddaf-Sh, M-Mahdi and Soltani, Iman},
  booktitle={2025 IEEE International Conference on Robotics and Automation (ICRA)},
  pages={7952--7959},
  year={2025},
  organization={IEEE}
}

@article{zhao2026cov,
  title={CoV: Chain-of-View Prompting for Spatial Reasoning},
  author={Zhao, Haoyu and Liu, Akide and Zhang, Zeyu and Wang, Weijie and Chen, Feng and Zhu, Ruihan and Haffari, Gholamreza and Zhuang, Bohan},
  journal={arXiv preprint arXiv:2601.05172},
  year={2026}
}

@article{zhang2026think3d,
  title={Think3D: Thinking with Space for Spatial Reasoning},
  author={Zhang, Zaibin and Wu, Yuhan and Jia, Lianjie and Wang, Yifan and Zhang, Zhongbo and Li, Yijiang and Ran, Binghao and Zhang, Fuxi and Sun, Zhuohan and Yin, Zhenfei and others},
  journal={arXiv preprint arXiv:2601.13029},
  year={2026}
}

@article{xiao2025ava,
  title={AVA-VLA: Improving Vision-Language-Action models with Active Visual Attention},
  author={Xiao, Lei and Li, Jifeng and Gao, Juntao and Ye, Feiyang and Jin, Yan and Qian, Jingjing and Zhang, Jing and Wu, Yong and Yu, Xiaoyuan},
  journal={arXiv preprint arXiv:2511.18960},
  year={2025}
}

@article{yu2025thinking,
  title={Thinking in 360 $\{$$\backslash$deg$\}$: Humanoid Visual Search in the Wild},
  author={Yu, Heyang and Han, Yinan and Zhang, Xiangyu and Yin, Baiqiao and Chang, Bowen and Han, Xiangyu and Liu, Xinhao and Zhang, Jing and Pavone, Marco and Feng, Chen and others},
  journal={arXiv preprint arXiv:2511.20351},
  year={2025}
}

@inproceedings{fu20213d,
  title={3d-front: 3d furnished rooms with layouts and semantics},
  author={Fu, Huan and Cai, Bowen and Gao, Lin and Zhang, Ling-Xiao and Wang, Jiaming and Li, Cao and Zeng, Qixun and Sun, Chengyue and Jia, Rongfei and Zhao, Binqiang and others},
  booktitle={Proceedings of the IEEE/CVF International Conference on Computer Vision},
  pages={10933--10942},
  year={2021}
}

@article{zhang2025spatial,
  title={Spatial understanding from videos: Structured prompts meet simulation data},
  author={Zhang, Haoyu and Liu, Meng and Li, Zaijing and Wen, Haokun and Guan, Weili and Wang, Yaowei and Nie, Liqiang},
  journal={arXiv preprint arXiv:2506.03642},
  year={2025}
}

@article{brown2025sims,
  title={Sims-v: Simulated instruction-tuning for spatial video understanding},
  author={Brown, Ellis and Ray, Arijit and Krishna, Ranjay and Girshick, Ross and Fergus, Rob and Xie, Saining},
  journal={arXiv preprint arXiv:2511.04668},
  year={2025}
}

@article{kolve2017ai2,
  title={Ai2-thor: An interactive 3d environment for visual ai},
  author={Kolve, Eric and Mottaghi, Roozbeh and Han, Winson and VanderBilt, Eli and Weihs, Luca and Herrasti, Alvaro and Deitke, Matt and Ehsani, Kiana and Gordon, Daniel and Zhu, Yuke and others},
  journal={arXiv preprint arXiv:1712.05474},
  year={2017}
}

@article{wang2024grutopia,
  title={Grutopia: Dream general robots in a city at scale},
  author={Wang, Hanqing and Chen, Jiahe and Huang, Wensi and Ben, Qingwei and Wang, Tai and Mi, Boyu and Huang, Tao and Zhao, Siheng and Chen, Yilun and Yang, Sizhe and others},
  journal={arXiv preprint arXiv:2407.10943},
  year={2024}
}

@inproceedings{li2023behavior,
  title={Behavior-1k: A benchmark for embodied ai with 1,000 everyday activities and realistic simulation},
  author={Li, Chengshu and Zhang, Ruohan and Wong, Josiah and Gokmen, Cem and Srivastava, Sanjana and Mart{\'\i}n-Mart{\'\i}n, Roberto and Wang, Chen and Levine, Gabrael and Lingelbach, Michael and Sun, Jiankai and others},
  booktitle={Conference on Robot Learning},
  pages={80--93},
  year={2023},
  organization={PMLR}
}

@inproceedings{dai2017scannet,
  title={Scannet: Richly-annotated 3d reconstructions of indoor scenes},
  author={Dai, Angela and Chang, Angel X and Savva, Manolis and Halber, Maciej and Funkhouser, Thomas and Nie{\ss}ner, Matthias},
  booktitle={Proceedings of the IEEE conference on computer vision and pattern recognition},
  pages={5828--5839},
  year={2017}
}

@inproceedings{yeshwanth2023scannet++,
  title={Scannet++: A high-fidelity dataset of 3d indoor scenes},
  author={Yeshwanth, Chandan and Liu, Yueh-Cheng and Nie{\ss}ner, Matthias and Dai, Angela},
  booktitle={Proceedings of the IEEE/CVF International Conference on Computer Vision},
  pages={12--22},
  year={2023}
}

@article{baruch2021arkitscenes,
  title={Arkitscenes: A diverse real-world dataset for 3d indoor scene understanding using mobile rgb-d data},
  author={Baruch, Gilad and Chen, Zhuoyuan and Dehghan, Afshin and Dimry, Tal and Feigin, Yuri and Fu, Peter and Gebauer, Thomas and Joffe, Brandon and Kurz, Daniel and Schwartz, Arik and others},
  journal={arXiv preprint arXiv:2111.08897},
  year={2021}
}

@article{team2023gemini,
  title={Gemini: a family of highly capable multimodal models},
  author={Team, Gemini and Anil, Rohan and Borgeaud, Sebastian and Alayrac, Jean-Baptiste and Yu, Jiahui and Soricut, Radu and Schalkwyk, Johan and Dai, Andrew M and Hauth, Anja and Millican, Katie and others},
  journal={arXiv preprint arXiv:2312.11805},
  year={2023}
}

@article{xu2025streamingvlm,
  title={Streamingvlm: Real-time understanding for infinite video streams},
  author={Xu, Ruyi and Xiao, Guangxuan and Chen, Yukang and He, Liuning and Peng, Kelly and Lu, Yao and Han, Song},
  journal={arXiv preprint arXiv:2510.09608},
  year={2025}
}

@article{zhuo2025streaming,
  title={Streaming 4d visual geometry transformer},
  author={Zhuo, Dong and Zheng, Wenzhao and Guo, Jiahe and Wu, Yuqi and Zhou, Jie and Lu, Jiwen},
  journal={arXiv preprint arXiv:2507.11539},
  year={2025}
}

@article{bai2025qwen3,
  title={Qwen3-vl technical report},
  author={Bai, Shuai and Cai, Yuxuan and Chen, Ruizhe and Chen, Keqin and Chen, Xionghui and Cheng, Zesen and Deng, Lianghao and Ding, Wei and Gao, Chang and Ge, Chunjiang and others},
  journal={arXiv preprint arXiv:2511.21631},
  year={2025}
}

@article{achiam2023gpt,
  title={Gpt-4 technical report},
  author={Achiam, Josh and Adler, Steven and Agarwal, Sandhini and Ahmad, Lama and Akkaya, Ilge and Aleman, Florencia Leoni and Almeida, Diogo and Altenschmidt, Janko and Altman, Sam and Anadkat, Shyamal and others},
  journal={arXiv preprint arXiv:2303.08774},
  year={2023}
}

@article{an2025llava,
  title={Llava-onevision-1.5: Fully open framework for democratized multimodal training},
  author={An, Xiang and Xie, Yin and Yang, Kaicheng and Zhang, Wenkang and Zhao, Xiuwei and Cheng, Zheng and Wang, Yirui and Xu, Songcen and Chen, Changrui and Zhu, Didi and others},
  journal={arXiv preprint arXiv:2509.23661},
  year={2025}
}

@article{zhang2024llava,
  title={Llava-video: Video instruction tuning with synthetic data},
  author={Zhang, Yuanhan and Wu, Jinming and Li, Wei and Li, Bo and Ma, Zejun and Liu, Ziwei and Li, Chunyuan},
  journal={arXiv preprint arXiv:2410.02713},
  year={2024}
}

@article{wang2025internvl3,
  title={Internvl3. 5: Advancing open-source multimodal models in versatility, reasoning, and efficiency},
  author={Wang, Weiyun and Gao, Zhangwei and Gu, Lixin and Pu, Hengjun and Cui, Long and Wei, Xingguang and Liu, Zhaoyang and Jing, Linglin and Ye, Shenglong and Shao, Jie and others},
  journal={arXiv preprint arXiv:2508.18265},
  year={2025}
}

@article{cai2025scaling,
  title={Scaling spatial intelligence with multimodal foundation models},
  author={Cai, Zhongang and Wang, Ruisi and Gu, Chenyang and Pu, Fanyi and Xu, Junxiang and Wang, Yubo and Yin, Wanqi and Yang, Zhitao and Wei, Chen and Sun, Qingping and others},
  journal={arXiv preprint arXiv:2511.13719},
  year={2025}
}

@article{yamada2023evaluating,
  title={Evaluating spatial understanding of large language models},
  author={Yamada, Yutaro and Bao, Yihan and Lampinen, Andrew K and Kasai, Jungo and Yildirim, Ilker},
  journal={arXiv preprint arXiv:2310.14540},
  year={2023}
}

@inproceedings{li2024mvbench,
  title={Mvbench: A comprehensive multi-modal video understanding benchmark},
  author={Li, Kunchang and Wang, Yali and He, Yinan and Li, Yizhuo and Wang, Yi and Liu, Yi and Wang, Zun and Xu, Jilan and Chen, Guo and Luo, Ping and others},
  booktitle={Proceedings of the IEEE/CVF Conference on Computer Vision and Pattern Recognition},
  pages={22195--22206},
  year={2024}
}

@article{xu2025spatialbench,
  title={Spatialbench: Benchmarking multimodal large language models for spatial cognition},
  author={Xu, Peiran and Wang, Sudong and Zhu, Yao and Li, Jianing and Zhang, Yunjian},
  journal={arXiv preprint arXiv:2511.21471},
  year={2025}
}

@inproceedings{tian2025nuscenes,
  title={Nuscenes-spatialqa: A spatial understanding and reasoning benchmark for vision-language models in autonomous driving},
  author={Tian, Kexin and Mao, Jingrui and Zhang, Yunlong and Jiang, Jiwan and Zhou, Yang and Tu, Zhengzhong},
  booktitle={Proceedings of the IEEE/CVF International Conference on Computer Vision},
  pages={4567--4576},
  year={2025}
}

@inproceedings{du2024embspatial,
  title={Embspatial-bench: Benchmarking spatial understanding for embodied tasks with large vision-language models},
  author={Du, Mengfei and Wu, Binhao and Li, Zejun and Huang, Xuan-Jing and Wei, Zhongyu},
  booktitle={Proceedings of the 62nd Annual Meeting of the Association for Computational Linguistics (Volume 2: Short Papers)},
  pages={346--355},
  year={2024}
}

@article{yu2025far,
  title={How far are vlms from visual spatial intelligence? a benchmark-driven perspective},
  author={Yu, Songsong and Chen, Yuxin and Ju, Hao and Jia, Lianjie and Zhang, Fuxi and Huang, Shaofei and Wu, Yuhan and Cui, Rundi and Ran, Binghao and Zhang, Zaibin and others},
  journal={arXiv preprint arXiv:2509.18905},
  year={2025}
}

@article{stogiannidis2025mind,
  title={Mind the gap: Benchmarking spatial reasoning in vision-language models},
  author={Stogiannidis, Ilias and McDonagh, Steven and Tsaftaris, Sotirios A},
  journal={arXiv preprint arXiv:2503.19707},
  year={2025}
}

@article{hu2025memory,
  title={Memory in the age of ai agents},
  author={Hu, Yuyang and Liu, Shichun and Yue, Yanwei and Zhang, Guibin and Liu, Boyang and Zhu, Fangyi and Lin, Jiahang and Guo, Honglin and Dou, Shihan and Xi, Zhiheng and others},
  journal={arXiv preprint arXiv:2512.13564},
  year={2025}
}

@article{zhang2025memory,
  title={Memory as action: Autonomous context curation for long-horizon agentic tasks},
  author={Zhang, Yuxiang and Shu, Jiangming and Ma, Ye and Lin, Xueyuan and Wu, Shangxi and Sang, Jitao},
  journal={arXiv preprint arXiv:2510.12635},
  year={2025}
}

@article{xu2025mem,
  title={A-mem: Agentic memory for llm agents},
  author={Xu, Wujiang and Liang, Zujie and Mei, Kai and Gao, Hang and Tan, Juntao and Zhang, Yongfeng},
  journal={arXiv preprint arXiv:2502.12110},
  year={2025}
}

@article{chhikara2025mem0,
  title={Mem0: Building production-ready ai agents with scalable long-term memory},
  author={Chhikara, Prateek and Khant, Dev and Aryan, Saket and Singh, Taranjeet and Yadav, Deshraj},
  journal={arXiv preprint arXiv:2504.19413},
  year={2025}
}

@article{yan2025memory,
  title={Memory-r1: Enhancing large language model agents to manage and utilize memories via reinforcement learning},
  author={Yan, Sikuan and Yang, Xiufeng and Huang, Zuchao and Nie, Ercong and Ding, Zifeng and Li, Zonggen and Ma, Xiaowen and Bi, Jinhe and Kersting, Kristian and Pan, Jeff Z and others},
  journal={arXiv preprint arXiv:2508.19828},
  year={2025}
}

@inproceedings{evans2024bad,
  title={Bad students make great teachers: Active learning accelerates large-scale visual understanding},
  author={Evans, Talfan and Pathak, Shreya and Merzic, Hamza and Schwarz, Jonathan and Tanno, Ryutaro and Henaff, Olivier J},
  booktitle={European Conference on Computer Vision},
  pages={264--280},
  year={2024},
  organization={Springer}
}

@inproceedings{song2025robospatial,
  title={Robospatial: Teaching spatial understanding to 2d and 3d vision-language models for robotics},
  author={Song, Chan Hee and Blukis, Valts and Tremblay, Jonathan and Tyree, Stephen and Su, Yu and Birchfield, Stan},
  booktitle={Proceedings of the Computer Vision and Pattern Recognition Conference},
  pages={15768--15780},
  year={2025}
}

@article{team2025robobrain,
  title={Robobrain 2.0 technical report},
  author={Team, BAAI RoboBrain and Cao, Mingyu and Tan, Huajie and Ji, Yuheng and Chen, Xiansheng and Lin, Minglan and Li, Zhiyu and Cao, Zhou and Wang, Pengwei and Zhou, Enshen and others},
  journal={arXiv preprint arXiv:2507.02029},
  year={2025}
}

@article{lin2024streamingbench,
  title={Streamingbench: Assessing the gap for mllms to achieve streaming video understanding},
  author={Lin, Junming and Fang, Zheng and Chen, Chi and Wan, Zihao and Luo, Fuwen and Li, Peng and Liu, Yang and Sun, Maosong},
  journal={arXiv preprint arXiv:2411.03628},
  year={2024}
}

@article{qian2024streaming,
  title={Streaming long video understanding with large language models},
  author={Qian, Rui and Dong, Xiaoyi and Zhang, Pan and Zang, Yuhang and Ding, Shuangrui and Lin, Dahua and Wang, Jiaqi},
  journal={Advances in Neural Information Processing Systems},
  volume={37},
  pages={119336--119360},
  year={2024}
}

@inproceedings{zhao2023streaming,
  title={Streaming video model},
  author={Zhao, Yucheng and Luo, Chong and Tang, Chuanxin and Chen, Dongdong and Codella, Noel and Zha, Zheng-Jun},
  booktitle={Proceedings of the IEEE/CVF conference on computer vision and pattern recognition},
  pages={14602--14612},
  year={2023}
}

@inproceedings{wang2025ross3d,
  title={Ross3d: Reconstructive visual instruction tuning with 3d-awareness},
  author={Wang, Haochen and Zhao, Yucheng and Wang, Tiancai and Fan, Haoqiang and Zhang, Xiangyu and Zhang, Zhaoxiang},
  booktitle={Proceedings of the IEEE/CVF International Conference on Computer Vision},
  pages={9275--9286},
  year={2025}
}

@article{li2025imagine,
  title={Imagine while reasoning in space: Multimodal visualization-of-thought},
  author={Li, Chengzu and Wu, Wenshan and Zhang, Huanyu and Xia, Yan and Mao, Shaoguang and Dong, Li and Vuli{\'c}, Ivan and Wei, Furu},
  journal={arXiv preprint arXiv:2501.07542},
  year={2025}
}

@inproceedings{lee2025perspective,
  title={Perspective-aware reasoning in vision-language models via mental imagery simulation},
  author={Lee, Phillip Y and Je, Jihyeon and Park, Chanho and Uy, Mikaela Angelina and Guibas, Leonidas and Sung, Minhyuk},
  booktitle={Proceedings of the IEEE/CVF International Conference on Computer Vision},
  pages={9241--9251},
  year={2025}
}

@article{yang2025mindjourney,
  title={MindJourney: Test-Time Scaling with World Models for Spatial Reasoning},
  author={Yang, Yuncong and Liu, Jiageng and Zhang, Zheyuan and Zhou, Siyuan and Tan, Reuben and Yang, Jianwei and Du, Yilun and Gan, Chuang},
  journal={arXiv preprint arXiv:2507.12508},
  year={2025}
}

@article{chen2026spatial,
  title={Spatial Chain-of-Thought: Bridging Understanding and Generation Models for Spatial Reasoning Generation},
  author={Chen, Wei and Long, Yancheng and Liu, Mingqiao and Ding, Haojie and Yang, Yankai and Wei, Hongyang and Zhang, Yi-Fan and Wen, Bin and Yang, Fan and Gao, Tingting and others},
  journal={arXiv preprint arXiv:2602.11980},
  year={2026}
}

@inproceedings{savva2019habitat,
  title={Habitat: A platform for embodied ai research},
  author={Savva, Manolis and Kadian, Abhishek and Maksymets, Oleksandr and Zhao, Yili and Wijmans, Erik and Jain, Bhavana and Straub, Julian and Liu, Jia and Koltun, Vladlen and Malik, Jitendra and others},
  booktitle={Proceedings of the IEEE/CVF international conference on computer vision},
  pages={9339--9347},
  year={2019}
}

@article{kerbl20233d,
  title={3d gaussian splatting for real-time radiance field rendering.},
  author={Kerbl, Bernhard and Kopanas, Georgios and Leimk{\"u}hler, Thomas and Drettakis, George and others},
  journal={ACM Trans. Graph.},
  volume={42},
  number={4},
  pages={139--1},
  year={2023}
}

\clearpage

\appendix

\section*{Appendix overview}
This document is organized as follows:

\begin{itemize}
    \item \hyperref[sec:details]{1. Details and discussion}

    \item \hyperref[sec:more_experiments]{2. More experiments}
    \begin{itemize}
        \item \hyperref[subsec:active_baselines]{2.1 Active Baselines}
        \item \hyperref[subsec:ablation_k]{2.2 Ablation on Memory Folding Interval $K$}
        \item \hyperref[subsec:data_scaling]{2.3 Data Scaling Analysis}
        \item \hyperref[subsec:blind_testing]{2.4 Blind Testing on $S^3$-Eval}
        \item \hyperref[subsec:detailed_results]{2.5 Detailed Results of Main Text Experiments}
    \end{itemize}
    
    \item \hyperref[sec:qa_logics]{3. QA generation logics}

    \item \hyperref[sec:visualization]{4. Visualization samples}
\end{itemize}

\phantomsection
\section*{1. Details and discussion} \label{sec:details}
\subsection*{1.1 Experimental Setup and details}
All training procedures were conducted using NVIDIA B200 GPUs, while the evaluation process, including latency measurements, was performed on NVIDIA H800 GPUs. All pretrained models are instruct version without thinking. For all testing phases, baselines lacking explicit temporal encoding or processing mechanisms were provided with time-related information directly described within the context. Additionally, the response formats for numerical and multiple-choice questions were predefined in the prompts. For our proposed method, the following system prompt was utilized: \textit{``You are a spatial understanding agent; based on the observed spatial information, answer the corresponding questions.''}

In the S$^3$-Bench dataset, all real-world video sequences are normalized to a resolution of $640 \times 480$ pixels with a frame rate of 24 frames per second (FPS), whereas the simulated data are standardized to a resolution of $768 \times 768$ pixels at 24 FPS. In terms of video duration, 25.3\% of the sequences are shorter than 30 seconds, 58.5\% range between 30 and 90 seconds, and the remaining 16\% exceed 90 seconds in length.

\subsection*{1.2 Data Construction Discussion} 
Regarding the data construction process, it is worth noting that some existing approaches rely on real-world scanned environments (e.g., Habitat \cite{savva2019habitat}) rather than simulated digital assets. However, the simulation realism and visual fidelity of these platforms are relatively limited. They are insufficient to support highly flexible exploration while maintaining stable rendering quality; therefore, they were not adopted in our study.

\section*{2. More experiments} \label{sec:more_experiments}
In this section, we provide supplementary experimental results, including comparisons with active vision baselines, an ablation study on the memory folding interval $K$, a data scaling analysis, comprehensive blind testing across all models on S$^3$-Eval, and detailed metrics for the experiments summarized in the main text.

\subsection*{2.1 Active Baselines} \label{subsec:active_baselines}
We incorporate CoV \cite{zhao2026cov} and Think3D \cite{zhang2026think3d} two spatial understanding methods based on active vision. Since these approaches are not inherently designed for online streaming comprehension, we present their evaluations in the appendix. As observed on the S$^3$-Eval benchmark (see Table \ref{tab:performance_sim_active} and Table \ref{tab:performance_real_active}), both methods exhibit limited performance. This can be primarily attributed to their constraints in acquiring precise novel information. Specifically, CoV relies on prompt engineering, which restricts the model's capacity for active exploration. Conversely, Think3D synthesizes novel views via 3DGS \cite{kerbl20233d} reconstruction; however, this mechanism primarily mines existing observations rather than introducing sufficiently new visual information.

\begin{table*}[t]
\centering
\caption{\small Performance comparison of active vision baselines across all $S^3$-Eval simulation part.}
\label{tab:performance_sim_active}
\tiny
\setlength{\tabcolsep}{1.35pt} 
\renewcommand{\arraystretch}{1.1} 

\begin{tabular}{l | *{17}{c}}
\toprule
Method & Over- & T-C & T- & T-Spa & Spa- & Spa- & C- & C- & C- & Id- & Rel- & Area & C- & Attr & Cou & Seq & Seq \\
 & all & -O-D & Hor & -D & D & Pro & O-D & Dist & Mo & Clo & Ori &  & Area &  &  &  & Id \\
\midrule
Qwen3-VL-8B \cite{bai2025qwen3} & 41.1 & 36.0 & 25.8 & 34.0 & 29.7 & 39.6 & 32.0 & 41.1 & 56.7 & 54.2 & 40.2 & 32.5 & 41.5 & 48.1 & 48.8 & 45.4 & 49.1 \\
Think3D \cite{zhang2026think3d} & 44.0 & 36.5 & 27.4 & 35.4 & 31.4 & 43.7 & 33.4 & 44.9 & 59.0 & 59.5 & 47.2 & 34.9 & 46.3 & 46.9 & 52.5 & 49.3 & 50.6 \\
CoV \cite{zhao2026cov} & 43.2 & 36.8 & 26.1 & 35.5 & 30.0 & 41.3 & 33.2 & 42.4 & 57.1 & 57.3 & 44.8 & 34.4 & 45.1 & 51.2 & 54.1 & 47.6 & 48.6 \\
\midrule
Ours & {62.9} & {57.6} & {35.0} & {55.6} & {59.4} & {70.4} & {70.8} & {67.7} & {75.6} & {72.8} & {46.3} & {67.0} & {68.1} & {70.8} & {61.8} & {60.9} & {60.6} \\
\bottomrule
\end{tabular}
\end{table*}

\begin{table*}[t]
\centering
\caption{\small Performance comparison of active vision baselines across all $S^3$-Eval real part.}
\label{tab:performance_real_active}
\tiny
\setlength{\tabcolsep}{1.35pt} 
\renewcommand{\arraystretch}{1.1} 

\begin{tabular}{l | *{16}{c}}
\toprule
Method & Over- & T-C & T- & T-Spa & Spa- & Spa- & C- & C- & C- & Id- & Rel- & Area & Attr & Cou & Seq & Seq \\
 & all & -O-D & Hor & -D & D & Pro & O-D & Dist & Mo & Clo & Ori &  &  &  &  & Id \\
\midrule
Qwen3-VL-8B \cite{bai2025qwen3} & 39.0 & 25.3 & 27.7 & 27.6 & 23.7 & 57.3 & 22.1 & 36.1 & 59.8 & 49.3 & 41.6 & 48.5 & 25.0 & 51.9 & 47.3 & 50.0 \\
Think3D \cite{zhang2026think3d} & 42.0 & 28.0 & 30.1 & 29.4 & 25.6 & 61.0 & 24.7 & 37.8 & 63.9 & 53.3 & 46.9 & 50.3 & 28.0 & 54.1 & 49.9 & 52.0 \\
CoV \cite{zhao2026cov} & 40.8 & 26.6 & 28.8 & 28.7 & 24.8 & 57.2 & 22.5 & 38.2 & 65.4 & 51.6 & 44.4 & 49.6 & 25.8 & 54.5 & 48.8 & 51.5 \\
\midrule
Ours & {57.8} & {41.0} & {37.0} & {48.2} & {50.1} & {71.8} & {45.1} & {77.5} & {75.5} & {59.7} & {44.5} & {49.7} & {70.9} & {64.0} & {72.0} & {64.4} \\
\bottomrule
\end{tabular}
\end{table*}

\subsection*{2.2 Ablation on Memory Folding Interval $K$} \label{subsec:ablation_k}
We investigate the impact of the memory folding interval, denoted as $K$, on the Sim split of S$^3$-Eval. As detailed in Table \ref{tab:ablation_k}, the empirical results indicate that an excessively small K leads to overly frequent summarization, generating redundant information that degrades overall performance. Conversely, an overly large $K$ fails to capture the scene dynamics comprehensively. Therefore, maintaining a moderate value for $K$ is essential for optimal performance.
\begin{table*}[t]
\centering
\caption{\small Ablation under different K.}
\label{tab:ablation_k}
\tiny
\setlength{\tabcolsep}{1.35pt} 
\renewcommand{\arraystretch}{1.1} 

\begin{tabular}{l | *{17}{c}}
\toprule
Method & Over- & T-C & T- & T-Spa & Spa- & Spa- & C- & C- & C- & Id- & Rel- & Area & C- & Attr & Cou & Seq & Seq \\
 & all & -O-D & Hor & -D & D & Pro & O-D & Dist & Mo & Clo & Ori &  & Area &  &  &  & Id \\
\midrule
K=2 & 60.2 & 53.2 & 33.8 & 51.3 & 57.9 & 68.5 & 66.7 & 65.3 & 69.1 & 69.5 & 45.1 & 65.7 & 64.5 & 69.7 & 58.9 & 59.2 & 58.8 \\
K=5 & 61.6 & 58.0 & 33.3 & 53.4 & 58.7 & 69.9 & 68.4 & 64.2 & 73.5 & 69.5 & 43.4 & 69.6 & 68.3 & 70.4 & 61.7 & 59.2 & 57.5 \\
{K=10} & {62.9} & {57.6} & {35.0} & {55.6} & {59.4} & {70.4} & {70.8} & {67.7} & {75.6} & {72.8} & {46.3} & {67.0} & {68.1} & {70.8} & {61.8} & {60.9} & {60.6} \\
K=15 & 62.4 & 56.8 & 34.5 & 50.8 & 59.7 & 70.9 & 73.8 & 68.2 & 72.8 & 69.7 & 45.9 & 64.8 & 69.1 & 69.7 & 61.7 & 61.1 & 61.9 \\
\bottomrule
\end{tabular}
\end{table*}

\subsection*{2.3 Data Scaling Analysis} \label{subsec:data_scaling}
Furthermore, we explore the effect of data scaling on the Sim split of S$^3$-Eval. As shown in Table \ref{tab:data_scaling}, by incrementally increasing the volume of the S$^3$-Train dataset from a small subset, we observe that the performance of our proposed method steadily improves, eventually approaching convergence as the data volume increases.
\begin{table*}[t]
\centering
\caption{\small Performance comparison under different data scale.}
\label{tab:data_scaling}
\tiny
\setlength{\tabcolsep}{1.35pt} 
\renewcommand{\arraystretch}{1.1} 

\begin{tabular}{l | *{17}{c}}
\toprule
Method & Over- & T-C & T- & T-Spa & Spa- & Spa- & C- & C- & C- & Id- & Rel- & Area & C- & Attr & Cou & Seq & Seq \\
 & all & -O-D & Hor & -D & D & Pro & O-D & Dist & Mo & Clo & Ori &  & Area &  &  &  & Id \\
\midrule
10K & 52.8 & 48.3 & 29.4 & 43.8 & 48.9 & 60.9 & 59.5 & 57.8 & 64.9 & 61.2 & 40.6 & 57.6 & 55.9 & 56.9 & 52.0 & 48.5 & 52.2 \\
100K & 58.4 & 53.9 & 33.3 & 52.6 & 52.3 & 65.5 & 66.5 & 63.0 & 68.6 & 70.2 & 43.0 & 63.2 & 61.3 & 65.3 & 55.0 & 58.3 & 56.3 \\
300K & 60.4 & 54.0 & 34.3 & 52.7 & 57.4 & 68.0 & 70.5 & 63.5 & 74.9 & 70.6 & 45.5 & 65.1 & 65.3 & 65.7 & 59.6 & 57.2 & 54.5 \\
{600K} & {62.9} & {57.6} & {35.0} & {55.6} & {59.4} & {70.4} & {70.8} & {67.7} & {75.6} & {72.8} & {46.3} & {67.0} & {68.1} & {70.8} & {61.8} & {60.9} & {60.6} \\
\bottomrule
\end{tabular}
\end{table*}

\subsection*{2.4 Blind Testing on S$^3$-Eval} \label{subsec:blind_testing}
To verify whether the models genuinely process visual information rather than over-relying on language priors, we conduct comprehensive blind tests on S$^3$-Eval, where visual inputs are intentionally omitted. The results, summarized in Table \ref{tab:blind_test_sim} and Table \ref{tab:blind_test_real}, demonstrate a significant performance degradation across all evaluated models, including those exposed to the training dataset. This confirms that the models fundamentally depend on visual observations to answer the posed questions, rather than simply overfitting to language priors.

\begin{table*}[t]
\centering
\caption{\small  Blind testing of baselines across all $S^3$-Eval simulation part. (*) indicates models also trained on our data. ($\dagger$) indicates streaming inference modification.}
\label{tab:blind_test_sim}
\tiny
\setlength{\tabcolsep}{1.35pt} 
\renewcommand{\arraystretch}{1.1} 

\begin{tabular}{l | *{18}{c}}
\toprule
Method & Over- & T-C & T- & T-Spa & Spa- & Spa- & C- & C- & C- & C- & Id- & Rel- & Area & C- & Attr & Cou & Seq & Seq \\
 & all & -O-D & Hor & -D & D & Pro & O-D & Rel & Dist & Mo & Clo & Ori &  & Area &  &  &  & Id \\
\midrule
GPT-5.2 \cite{achiam2023gpt} & 26.0 & 7.8 & 31.2 & 35.8 & 39.7 & 33.7 & 26.2 & 31.1 & 20.0 & 38.6 & 28.8 & 48.4 & 7.5 & 28.6 & 1.8 & 1.0 & 34.9 & 35.0 \\
Gemini-3-flash \cite{team2023gemini} & 27.1 & 8.2 & 33.0 & 38.7 & 43.2 & 36.4 & 26.5 & 30.2 & 22.1 & 40.3 & 29.8 & 47.4 & 7.3 & 30.6 & 1.8 & 1.0 & 37.2 & 35.0 \\
Gemini-3-pro \cite{team2023gemini} & 26.8 & 8.1 & 31.2 & 36.6 & 42.0 & 35.7 & 26.3 & 32.6 & 21.9 & 41.0 & 28.7 & 47.6 & 7.4 & 29.7 & 1.8 & 1.0 & 34.7 & 37.9 \\

\midrule

llava-video \cite{zhang2024llava} & 21.0 & 6.4 & 24.8 & 28.7 & 33.5 & 28.2 & 20.9 & 24.4 & 16.9 & 30.0 & 22.7 & 37.5 & 6.3 & 23.4 & 1.3 & 0.8 & 27.3 & 29.7 \\
llava-ov-1.5-4B \cite{an2025llava} & 19.7 & 5.7 & 23.0 & 25.6 & 29.3 & 26.0 & 20.1 & 22.9 & 14.9 & 30.4 & 20.9 & 35.6 & 5.7 & 22.8 & 1.3 & 0.8 & 27.6 & 28.1 \\
llava-ov-1.5-8B \cite{an2025llava} & 23.7 & 6.6 & 27.6 & 34.6 & 38.5 & 32.3 & 22.8 & 27.6 & 18.0 & 34.7 & 25.7 & 40.6 & 6.7 & 27.3 & 1.5 & 0.9 & 33.5 & 31.2 \\

Qwen3-VL-4B \cite{bai2025qwen3} & 21.3 & 0.2 & 26.1 & 26.2 & 34.2 & 26.9 & 33.3 & 23.2 & 20.0 & 33.5 & 22.4 & 32.0 & 3.5 & 28.7 & 3.9 & 0.0 & 25.6 & 28.1 \\
Qwen3-VL-8B \cite{bai2025qwen3} & 15.6 & 0.2 & 22.0 & 23.8 & 27.0 & 26.0 & 0.7 & 25.1 & 0.3 & 24.8 & 23.5 & 37.5 & 0.0 & 10.0 & 0.0 & 1.6 & 28.7 & 24.6 \\
Qwen3-VL-32B \cite{bai2025qwen3} & 25.5 & 18.5 & 26.6 & 34.9 & 38.2 & 29.1 & 28.3 & 25.1 & 28.6 & 36.3 & 21.1 & 40.0 & 14.3 & 33.0 & 0.2 & 0.8 & 31.3 & 32.5 \\

InternVL3.5-1B \cite{wang2025internvl3} & 13.1 & 0.0 & 25.8 & 4.1 & 5.2 & 23.3 & 0.1 & 22.1 & 0.0 & 21.8 & 17.2 & 44.4 & 0.0 & 0.0 & 19.1 & 1.2 & 26.2 & 20.2 \\
InternVL3.5-8B \cite{wang2025internvl3} & 15.7 & 0.0 & 24.3 & 0.9 & 0.7 & 26.2 & 0.2 & 26.0 & 18.1 & 33.3 & 24.0 & 43.2 & 0.0 & 0.0 & 8.0 & 8.7 & 28.7 & 29.9 \\
InternVL3.5-38B \cite{wang2025internvl3} & 25.8 & 7.6 & 31.7 & 29.0 & 21.3 & 33.7 & 12.7 & 29.8 & 34.8 & 30.2 & 29.2 & 44.0 & 10.6 & 0.0 & 17.8 & 52.1 & 26.8 & 29.9 \\

\midrule

vlm-3r \cite{fan2025vlm} & 29.9 & 8.5 & 33.8 & 41.9 & 46.8 & 37.6 & 32.6 & 34.8 & 23.8 & 47.5 & 31.1 & 53.7 & 8.2 & 32.3 & 1.8 & 1.2 & 40.6 & 41.7 \\
cambrian-s \cite{yang2025cambrian} & 28.7 & 8.6 & 35.3 & 39.3 & 43.9 & 37.4 & 28.3 & 33.8 & 22.9 & 43.3 & 30.5 & 50.1 & 8.0 & 33.7 & 1.9 & 1.1 & 39.2 & 39.4 \\
\midrule
cambrian-s*${\dagger}$ \cite{yang2025cambrian} & 35.4 & 30.0 & 41.7 & 36.5 & 36.3 & 40.5 & 33.9 & 34.0 & 44.1 & 46.4 & 39.0 & 37.3 & 29.2 & 34.2 & 22.8 & 33.1 & 41.4 & 21.5 \\
streaming-vlm* \cite{xu2025streamingvlm} & 33.8 & 27.8 & 37.9 & 35.0 & 35.1 & 40.1 & 32.4 & 30.0 & 42.3 & 43.3 & 38.1 & 36.6 & 27.5 & 31.4 & 22.2 & 30.7 & 42.8 & 21.2 \\

{Ours} & {36.3} & {29.5} & {42.5} & {36.4} & {36.4} & {43.5} & {35.1} & {36.7} & {42.9} & {44.7} & {42.1} & {39.8} & {29.6} & {34.3} & {23.5} & {32.8} & {43.4} & {23.0} \\
\bottomrule
\end{tabular}
\end{table*}

\begin{table*}[t]
\centering
\caption{\small Blind testing of baselines across all $S^3$-Eval real part. (*) indicates models also trained on our data. ($\dagger$) indicates streaming inference modification.}
\label{tab:blind_test_real}
\tiny
\setlength{\tabcolsep}{1.35pt} 
\renewcommand{\arraystretch}{1.1} 

\begin{tabular}{l | *{17}{c}}
\toprule
Method & Over- & T-C & T- & T-Spa & Spa- & Spa- & C- & C- & C- & C- & Id- & Rel- & Area & Attr & Cou & Seq & Seq \\
 & all & -O-D & Hor & -D & D & Pro & O-D & Rel & Dist & Mo & Clo & Ori &  &  &  &  & Id \\
\midrule
GPT-5.2 \cite{achiam2023gpt} & 26.5 & 8.9 & 28.2 & 32.0 & 36.1 & 43.1 & 24.1 & 34.3 & 6.3 & 25.8 & 36.2 & 45.7 & 7.3 & 2.0 & 0.2 & 32.0 & 34.2 \\
Gemini-3-flash \cite{team2023gemini} & 28.3 & 9.0 & 28.6 & 34.3 & 40.6 & 45.7 & 27.3 & 36.4 & 6.8 & 28.1 & 38.1 & 49.9 & 7.6 & 2.1 & 0.2 & 33.8 & 35.1 \\
Gemini-3-pro \cite{team2023gemini} & 27.6 & 9.0 & 27.8 & 36.4 & 37.3 & 46.4 & 27.2 & 35.4 & 6.6 & 27.1 & 35.8 & 44.2 & 7.2 & 2.1 & 0.2 & 35.0 & 35.5 \\

\midrule
llava-video \cite{zhang2024llava} & 27.5 & 9.0 & 30.0 & 32.6 & 40.4 & 46.4 & 24.7 & 35.9 & 6.6 & 27.4 & 36.1 & 45.9 & 6.9 & 2.0 & 0.2 & 33.5 & 35.8 \\
llava-ov-1.5-4B \cite{an2025llava} & 26.5 & 8.7 & 27.9 & 33.7 & 37.9 & 44.4 & 26.2 & 34.6 & 6.3 & 27.0 & 34.6 & 43.0 & 6.9 & 2.0 & 0.2 & 30.8 & 33.3 \\
llava-ov-1.5-8B \cite{an2025llava} & 32.1 & 9.7 & 33.3 & 41.4 & 44.2 & 54.1 & 29.4 & 40.2 & 7.7 & 31.8 & 42.3 & 54.7 & 8.5 & 2.4 & 0.2 & 38.3 & 40.8 \\

Qwen3-VL-4B \cite{bai2025qwen3} & 23.0 & 0.3 & 25.1 & 31.8 & 38.1 & 37.7 & 31.7 & 28.1 & 9.9 & 23.6 & 28.3 & 29.3 & 0.0 & 4.6 & 0.0 & 27.3 & 27.1 \\
Qwen3-VL-8B \cite{bai2025qwen3} & 18.8 & 0.2 & 20.0 & 21.2 & 17.1 & 37.9 & 1.7 & 28.0 & 1.0 & 20.0 & 30.0 & 38.6 & 0.0 & 0.0 & 0.2 & 26.7 & 31.0 \\
Qwen3-VL-32B \cite{bai2025qwen3} & 25.5 & 21.3 & 25.6 & 31.4 & 39.0 & 38.7 & 29.9 & 30.1 & 5.2 & 21.9 & 30.3 & 43.3 & 17.5 & 0.3 & 0.3 & 29.3 & 28.3 \\

InternVL3.5-1B \cite{wang2025internvl3} & 19.4 & 0.0 & 37.0 & 3.8 & 10.1 & 38.9 & 0.3 & 26.4 & 0.9 & 28.6 & 25.4 & 45.9 & 0.0 & 23.9 & 1.8 & 29.0 & 24.5 \\
InternVL3.5-8B \cite{wang2025internvl3} & 18.7 & 0.0 & 22.0 & 0.4 & 6.4 & 39.0 & 5.2 & 26.2 & 0.9 & 25.5 & 28.7 & 46.5 & 0.0 & 6.5 & 14.9 & 28.8 & 32.5 \\
InternVL3.5-38B \cite{wang2025internvl3} & 29.1 & 18.2 & 29.3 & 30.8 & 34.5 & 37.7 & 22.8 & 27.8 & 4.8 & 21.5 & 27.4 & 47.0 & 33.2 & 9.6 & 53.2 & 32.1 & 34.7 \\

\midrule
vlm-3r \cite{fan2025vlm} & 27.5 & 8.6 & 29.9 & 34.4 & 39.6 & 43.7 & 26.9 & 33.2 & 6.5 & 26.7 & 36.1 & 46.2 & 7.1 & 2.0 & 0.2 & 32.7 & 37.3 \\
cambrian-s \cite{yang2025cambrian} & 28.8 & 9.5 & 30.5 & 35.6 & 41.1 & 49.9 & 27.2 & 38.6 & 6.9 & 28.4 & 36.9 & 46.7 & 7.7 & 2.1 & 0.2 & 34.2 & 37.8 \\
\midrule
cambrian-s*$^{\dagger}$ \cite{yang2025cambrian} & 33.5 & 29.8 & 38.1 & 33.3 & 35.8 & 41.6 & 30.8 & 33.6 & 30.5 & 35.5 & 39.2 & 37.9 & 28.2 & 23.8 & 31.6 & 39.7 & 18.4 \\
streaming-vlm* \cite{xu2025streamingvlm} & 34.9 & 30.1 & 39.3 & 33.4 & 37.3 & 43.4 & 32.4 & 35.5 & 31.2 & 36.3 & 41.8 & 40.2 & 29.7 & 24.6 & 33.3 & 41.9 & 19.8 \\
{Ours} & {33.2} & {29.8} & {38.2} & {30.8} & {36.1} & {41.2} & {31.5} & {34.2} & {30.4} & {37.3} & {40.1} & {37.7} & {26.7} & {24.1} & {29.7} & {37.7} & {16.9} \\
\bottomrule
\end{tabular}
\end{table*}

\subsection*{2.5 Detailed Results of Main Text Experiments} \label{subsec:detailed_results}
Finally, we present the extended quantitative results for the experiments briefly summarized in the main text. This includes comprehensive metrics for the module ablation studies (Table \ref{tab:ablation_details}), as well as the detailed performance breakdowns on other relevant benchmarks (Tables \ref{tab:vsi_details}, \ref{tab:blink_details}, and \ref{tab:mmsi_details}).
\begin{table*}[t]
\centering
\caption{\small Detailed module ablation study.}
\label{tab:ablation_details}
\tiny
\setlength{\tabcolsep}{1.35pt} 
\renewcommand{\arraystretch}{1.1} 

\begin{tabular}{l | *{17}{c}}
\toprule
Method & Over- & T-C & T- & T-Spa & Spa- & Spa- & C- & C- & C- & Id- & Rel- & Area & C- & Attr & Cou & Seq & Seq \\
 & all & -O-D & Hor & -D & D & Pro & O-D & Dist & Mo & Clo & Ori &  & Area &  &  &  & Id \\
\midrule
Baseline (data) & 56.0 & 54.9 & 31.0 & 51.8 & 45.2 & 58.3 & 66.0 & 60.9 & 67.0 & 65.6 & 40.0 & 59.1 & 60.1 & 63.2 & 55.8 & 55.2 & 55.0 \\
\quad + active & 58.1 & 54.9 & 31.0 & 51.8 & 53.0 & 64.8 & 66.0 & 60.9 & 67.0 & 65.6 & 43.6 & 63.4 & 64.5 & 67.4 & 59.2 & 55.2 & 55.0 \\
\quad + folding & 60.9 & 57.2 & 34.5 & 55.0 & 58.4 & 69.7 & 67.0 & 62.2 & 68.9 & 67.1 & 45.9 & 66.4 & 65.2 & 70.3 & 60.9 & 60.0 & 59.9 \\
\quad + 3D enc & {62.9} & {57.6} & {35.0} & {55.6} & {59.4} & {70.4} & {70.8} & {67.7} & {75.6} & {72.8} & {46.3} & {67.0} & {68.1} & {70.8} & {61.8} & {60.9} & {60.6} \\
\bottomrule
\end{tabular}
\end{table*}

\begin{table*}[t]
\centering
\caption{\small Detailed results on VSI-Bench. (*) denotes that the model is not trained on our data.}
\label{tab:vsi_details}
\tiny
\setlength{\tabcolsep}{1.35pt} 
\renewcommand{\arraystretch}{1.1} 

\begin{tabular}{l | *{11}{c}}
\toprule
Method & Over- & App- & Abs- & Obj- & Dir- & Dir- & Dir- & Rel- & Obj- & Room- & Route \\
 & all & Order & Dist & Count & Easy & Hard & Med & Dist & Size & Size & \\
\midrule

Ours* & 55.0 & 61.2 & 38.1 & 57.2 & 61.3 & 61.1 & 61.1 & 61.0 & 58.8 & 53.5 & 33.0 \\
Ours & 67.5 & 75.1 & 46.8 & 70.2 & 75.1 & 75.1 & 75.1 & 74.9 & 72.2 & 65.7 & 40.2 \\

\bottomrule
\end{tabular}
\end{table*}

\begin{table*}[t]
\centering
\caption{\small Detailed results on BLINK. (*) denotes that the model is not trained on our data.}
\label{tab:blink_details}
\tiny
\setlength{\tabcolsep}{1.35pt} 
\renewcommand{\arraystretch}{1.1} 

\begin{tabular}{l | *{15}{c}}
\toprule
Method & Over- & Sim- & Count- & Depth & Jigsaw & Art & Fun. & Sem. & Spatial & Local. & Vis. & Multi- & Reflect. & Foren- & IQ \\
 & all & ilarity & ing &  &  &  & Corr. & Corr. &  &  & Corr. & view &  & sic &  \\
\midrule

Ours* & 62.4 & 88.2 & 75.0 & 87.1 & 62.0 & 75.2 & 33.8 & 47.9 & 88.1 & 67.2 & 80.8 & 42.9 & 30.6 & 71.2 & 26.0 \\
Ours & 65.2 & 92.6 & 78.3 & 90.3 & 64.7 & 78.6 & 35.4 & 49.3 & 91.6 & 70.4 & 84.9 & 45.1 & 32.1 & 74.2 & 26.7 \\

\bottomrule
\end{tabular}
\end{table*}

\begin{table*}[t]
\centering
\caption{\small Detailed results on MMSI-Bench. (*) denotes that the model is not trained on our data.}
\label{tab:mmsi_details}
\tiny
\setlength{\tabcolsep}{1.35pt} 
\renewcommand{\arraystretch}{1.1} 

\begin{tabular}{l | *{12}{c}}
\toprule
Method & Over- & Cam.- & Obj.- & Reg.- & Cam.- & Obj.- & Reg.- & Meas. & Appr. & Cam. & Obj. & MSR \\
 & all & Cam. & Obj. & Reg. & Obj. & Reg. & Cam. &  &  &  &  &  \\

\midrule
Ours* & 31.2 & 29.0 & 29.8 & 29.6 & 31.4 & 35.3 & 31.3 & 29.7 & 21.2 & 24.3 & 47.4 & 31.8 \\
Ours & 30.5 & 28.0 & 28.7 & 28.4 & 30.2 & 34.1 & 30.1 & 29.7 & 21.2 & 24.3 & 46.1 & 31.8 \\
\bottomrule
\end{tabular}
\end{table*}

\section*{3. QA Generation Logics} \label{sec:qa_logics}
In this section, we detail the logical framework and construction pipeline for all Question-Answer (QA) pairs. The generation process fundamentally relies on the following initial inputs: the video sequences, corresponding camera extrinsics, 3D bounding box (bbox) annotations of objects within the scene, and per-frame object visibility.

The acquisition of object visibility varies depending on the data source. For simulated environments, visibility states are directly extracted from the rendering pipeline.  Conversely, for real-world datasets, visibility is determined through projection and occlusion calculations utilizing 3D annotations, such as point clouds. All subsequent generation algorithms and logical deductions are executed based on these foundational elements.

\subsection*{3.1 Single-Turn Question Generation Logic}
\begin{itemize}
    \item \textbf{Count (Visible Quantity Statistics):} Iterates through historical frames up to the current timestamp to accumulate the occurrence frequency of each object category. Categories with an instance count $> 1$ are prioritized for question formulation (e.g., ``How many chairs appeared?''). If none exist, categories with a single instance are utilized as fallbacks.
    \item \textbf{Attribute (Maximum Bounding Box Edge):} Randomly selects one unique object observed thus far. Extracts its pre-annotated 3D bounding box and returns the maximum dimension among its length, width, and height as the correct answer.
    \item \textbf{Sequence (Chronological Ordering):} Randomly samples three previously observed objects. Traces their initial appearance frame indices and sorts them chronologically (e.g., $A \rightarrow B \rightarrow C$). Incorrect permutations are automatically generated as distractors for multiple-choice formats.
    \item \textbf{Spatial Distance (Absolute Inter-object Distance):} Filters two previously observed objects located within the same room. Calculates the Euclidean distance between their 3D coordinates.
    \item \textbf{Area (Total Visited Area):} Determines the rooms traversed based on the camera trajectory. Aggregates the area of all rooms that satisfy a predefined visit ratio threshold (\texttt{AREA\_VISIT\_THRESHOLD}).
    \item \textbf{Spatial Proximity (Nearest Object in Space):} Randomly selects an anchor object. Calculates and sorts the Euclidean distances from all other unique objects in the field of view to this anchor. The nearest object is assigned as the ground truth; distant or unobserved categories serve as distractors.
    \item \textbf{Relative Orientation:} Selects three objects ($A$, $B$, $C$) within the same room. Assuming an observer at $A$ faces $B$ (forming the forward direction vector $\vec{AB}$), the relative position of $C$ is determined using the dot product and cross product. An angle exceeding $135^\circ$ is classified as ``Back''; otherwise, the sign of the cross product determines ``Left'' or ``Right''.
    \item \textbf{Camera Displacement:} Identifies frames indicating substantive forward movement (\texttt{MOVE\_TO\_NEXT}). Traces the camera's coordinates over the preceding $1$ to $3$ seconds to calculate the actual Euclidean displacement within this temporal window.
    \item \textbf{Camera-to-Object Absolute Distance:} Selects a valid, visible object during an ongoing action. Computes the direct Euclidean distance between the object's 3D coordinates and the current camera position.
    \item \textbf{Camera-to-Object Relative Distance:} Compiles all unique visible objects in the current frame and calculates their respective distances to the camera. Samples $2$ to $4$ objects as candidates, prompting the model to identify the one closest to the camera.
    \item \textbf{Current Room Area:} Extracts the camera's 3D coordinates in the current frame to reference the enclosing room boundaries (Hull Vertices), returning the predefined area of the specified room.
\end{itemize}

\subsection*{3.2 Multi-Turn Dependency Question Generation Logic}
\begin{enumerate}
    \item \textbf{Chronological Dependency}
    \begin{itemize}
        \item \textit{Q1 (sequence\_identification):} Identifies which object appeared earliest among a given set, utilizing the first-appearance frame logic.
        \item \textit{Q2 (tem\_horizontal\_direction):} Advances the timeline by a set number of frames. Queries the current on-screen region (e.g., front-left, front-right) of the target object identified in Q1. This is calculated by comparing the camera's 2D yaw angle with the object's actual azimuth.
    \end{itemize}
    \item \textbf{Spatial Superlative Dependency}
    \begin{itemize}
        \item \textit{Q1 (identification\_closest):} Presents multiple candidates and queries which object is currently closest to the camera.
        \item \textit{Q2 (tem\_spatial\_distance\_ref):} Advances the timeline and introduces a new reference object. Queries the distance between the previously identified closest object (from Q1) and the new reference.
    \end{itemize}
    \item \textbf{Camera Motion Dependency}
    \begin{itemize}
        \item \textit{Q1 (camera\_motion\_target):} Calculates the distance variation ($\Delta$) between visible objects and the camera over the preceding $3$ seconds. The object exhibiting the maximum distance reduction (largest negative $\Delta$) is identified as the target the camera is approaching.
        \item \textit{Q2 (tem\_cam\_obj\_distance\_ref):} Advances to a future frame and queries the remaining distance between the camera and the target object identified in Q1.
    \end{itemize}

\end{enumerate}

\section*{4. Visualization Samples} \label{sec:visualization}
\begin{figure}
    \centering
    \includegraphics[width=0.95\linewidth]{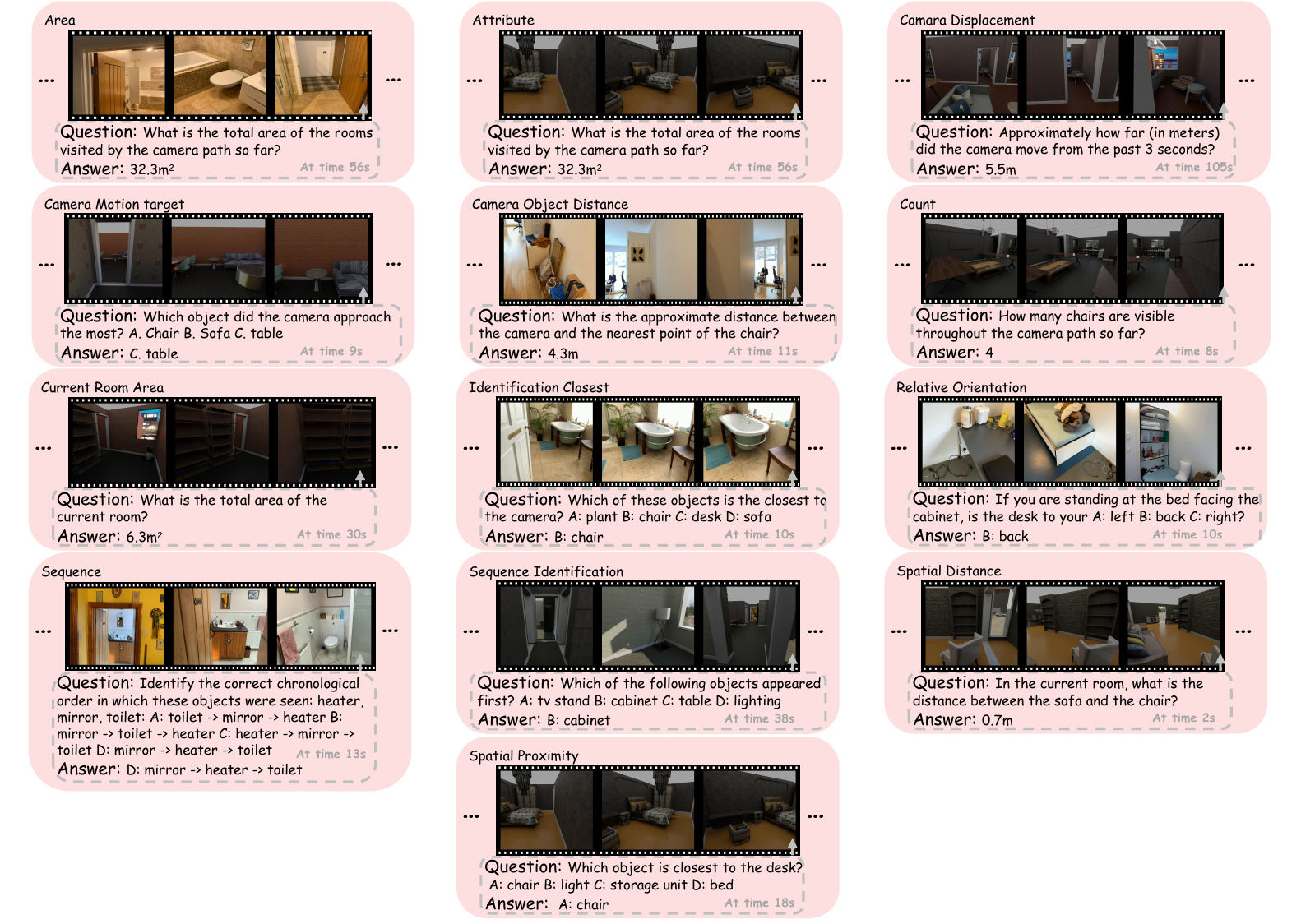}
    \caption{Single round QA pair samples of $S^3$-Bench.}
    \label{fig:samples_1}
\end{figure}

\begin{figure}
    \centering
    \includegraphics[width=0.95\linewidth]{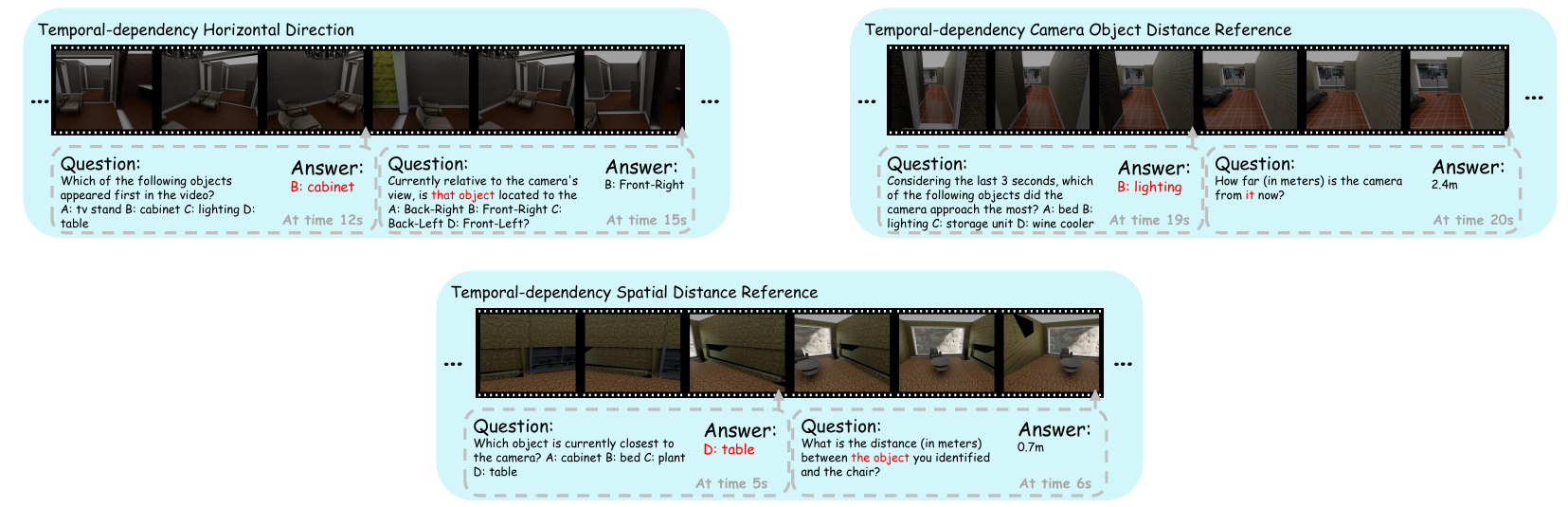}
    \caption{Temporal dependency QA pair samples of $S^3$-Bench.}
    \label{fig:samples_2}
\end{figure}

\begin{figure}
    \centering
    \includegraphics[width=0.95\linewidth]{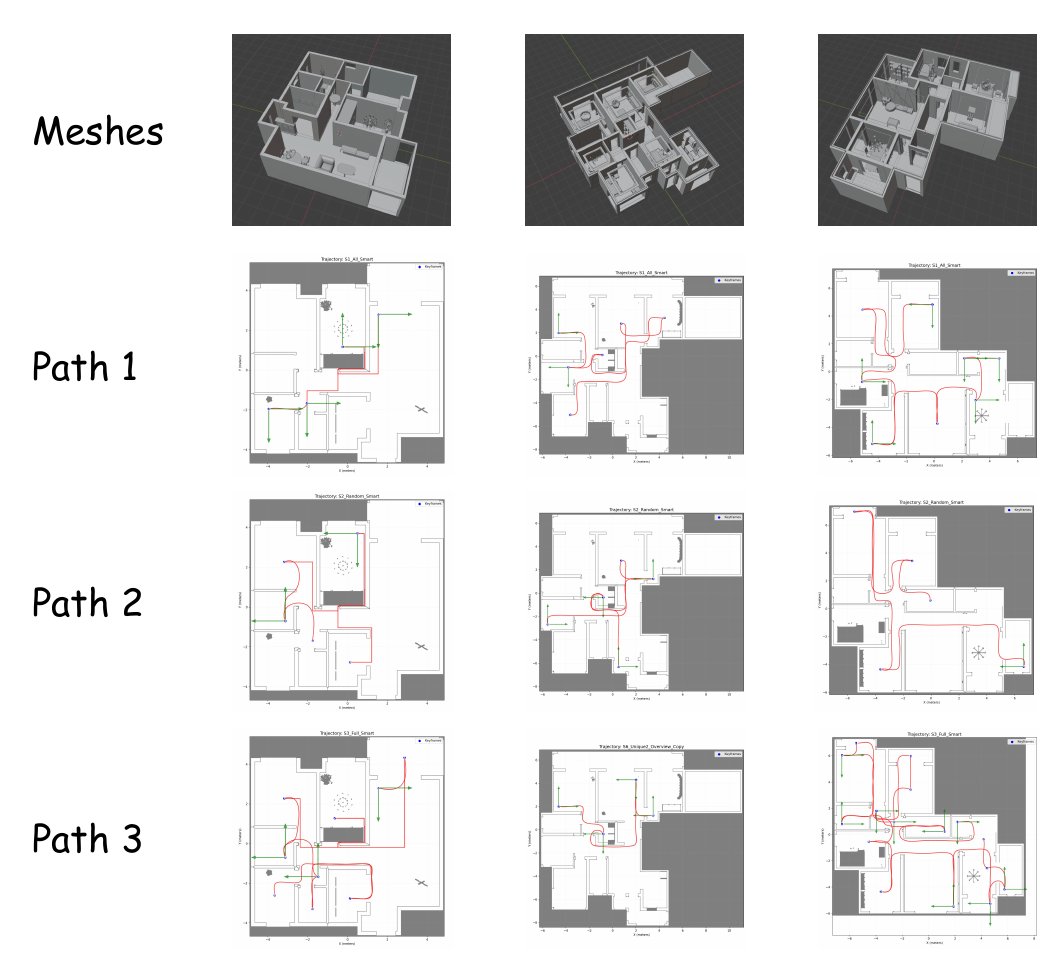}
    \caption{Samples of meshes of the simulation part of $S^3$-Bench, with the related diverse planned trajectories.}
    \label{fig:mesh_traj}
\end{figure}

In this section, we present comprehensive visualization examples of the generated Question-Answer (QA) pairs across various tasks in Fig.\ref{fig:samples_1},\ref{fig:samples_2}. These qualitative results illustrate the contextual grounding of the QA pairs within the environments, explicitly demonstrating the correspondence between the formulated questions, their respective answers, and the spatial configurations of the scenes. There are also some visual samples of 3D meshes and related diverse planned trejectories in Fig.\ref{fig:mesh_traj}. Our data construction pipeline generates various smooth trajectories to enhance the data diversity. Please refer to the attached multimedia files for video samples demonstrating the streaming online understanding processes.

\end{document}